\pdfoutput=1

\documentclass[11pt]{article}

\usepackage[final]{./acl}

\usepackage{times}
\usepackage{latexsym}
\usepackage{amsmath}
\usepackage{amssymb}
\usepackage{subcaption}
\usepackage{enumitem}
\usepackage[T1]{fontenc}

\usepackage[utf8]{inputenc}

\usepackage{microtype}

\usepackage{inconsolata}

\usepackage{graphicx}
\usepackage{float}
\graphicspath{{figs/}}

\title{
    Distribution Prompting: Understanding the Expressivity of Language Models
    Through the Next-Token Distributions They Can Produce
}

\author{Haojin Wang\textsuperscript{1}\thanks{Now at the University of Illinois Urbana-Champaign. Work done during visiting term at the University of Waterloo.} \quad Zining Zhu \textsuperscript{2}\quad Freda Shi \textsuperscript{1, 3} \\
                \textsuperscript{1} University of Waterloo\quad \textsuperscript{2} Stevens Institute of Technology \\
                \textsuperscript{3}
                Vector Institute, Canada CIFAR AI Chair
                \\\texttt{\{s2323wang,fhs\}@uwaterloo.ca}\quad
                \texttt{zzhu41@stevens.edu}
                }

\usepackage{hyperref}
\usepackage{cleveref}

\crefname{section}{\S\!}{\S\S\!}
\crefname{table}{Tab.}{Tabs.}
\crefname{figure}{Fig.}{Figs.}
\crefname{algorithm}{Alg.}{Algs.}
\crefname{appendix}{App.}{Apps.}
\crefname{equation}{Eq.}{Eqs.}

\newcommand{\interalia}[1]{\citep[\textit{inter alia}]{#1}}

\begin{document}
\maketitle
\begin{abstract}
    Autoregressive neural language models (LMs) generate a probability distribution over tokens at each time step given a prompt.
    In this work, we attempt to systematically understand the probability distributions that LMs can produce, showing that some distributions are significantly harder to elicit than others.
    Specifically, for any target next-token distribution over the vocabulary, we attempt to find a prompt that induces the LM to output a distribution as close as possible to the target, using either soft \citep{li_prefix-tuning_2021} or hard \citep{wallace2021universaladversarialtriggersattacking} gradient-based prompt tuning.
    We find that (1) in general, distributions with very low or very high entropy are easier to approximate than those with moderate entropy; (2) among distributions with the same entropy, those containing ``outlier tokens'' are easier to approximate; (3) target distributions generated by LMs---even LMs with different tokenizers---are easier to approximate than randomly chosen targets.
    These results offer insights into the expressiveness of LMs and the challenges of using them as probability distribution proposers.
\end{abstract}
\section{Introduction}
Pretrained language models (LMs) have acquired extensive knowledge \citep{zhao2024surveylargelanguagemodels} and shown strong performance across various tasks \citep{kaplan2020scalinglawsneurallanguage,DBLP:journals/corr/abs-2005-14165}.
Their impressive generative capabilities rely heavily on algorithms that determine the next token \interalia{finlayson2023closingcuriouscaseneural}.
As a foundation, these algorithms typically involve the softmax operation \citep{NIPS1989_0336dcba}, which computes the probability distribution of next tokens using the product of hidden states and token embeddings.
However, such operation leads to the softmax bottleneck \citep{yang2018breakingsoftmaxbottleneckhighrank},
which restricts the output distribution to a low-dimensional subspace, thereby limiting LM capability to express certain probability distributions.
For example, \citet{chang2022softmax} found that when predicting the next-word probabilities given ambiguous contexts, GPT-2 \citep{radford2019language} is often incapable of assigning the highest probabilities to nonsynonymous candidates.
In another line, \citet{finlayson2024logitsapiprotectedllmsleak} found that by leveraging the limitations in LM expressiveness, we can efficiently extract proprietary information about API-protected LMs.
These findings naturally raise a research question: are there general properties of the next-token probability distribution that can be easily elicited from LMs through prompting?

\begin{figure}[t]
    \includegraphics[width=\linewidth]{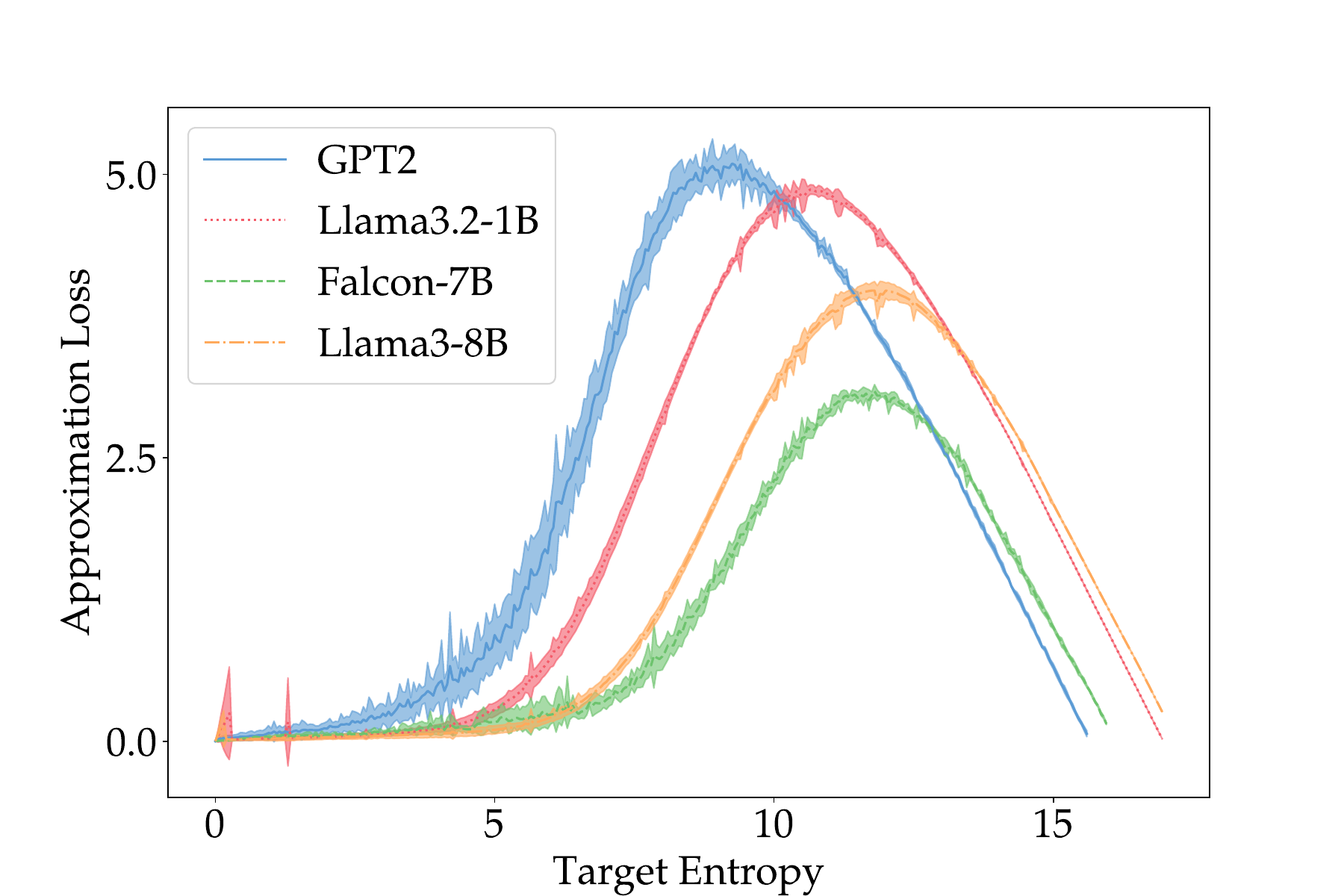}
    \vspace{-15pt}
    \caption{
        The Kullback--Leibler divergence between the target and approximated distributions for different entropy levels (in bits), using soft prompt tuning \citep{li_prefix-tuning_2021} with different pretrained LMs as the backbone.
        The shaded regions represent the standard error bands.
    }
    \vspace{-5pt}
    \label{Intro}
\end{figure}

To answer this question, we conduct a systematic analysis of the expressiveness of LMs, in terms of approximating different target probability distributions.
Without changing the parameters of LMs, we aim to find prompts that lead to the output of a probability distribution as close as possible to a target distribution. 
This design isolates the intrinsic expressivity of pretrained LMs, since we probe their behavior by varying only the prompts while keeping model parameters fixed.
Apparently, due to inefficiency for enumerating the infinitely many prompts as triggers for approximating the target distribution, we use modern optimization techniques. 
Through either soft \citep{li_prefix-tuning_2021} or hard \citep{wallace2021universaladversarialtriggersattacking}\footnote{The key difference between soft and hard prompts is that soft prompts can be continuous embeddings that do not align to any token in the vocabulary, while hard prompts must correspond to vocabulary entries.} prompt tuning across different initializations, we measure the difficulty of an approximation using the minimum Kullback--Leibler (KL) divergence between the target and the approximated distributions from different prompt initializations.

We conduct experiments on multiple pretrained LMs, including GPT-2 \citep{radford2019language}, Llama3.2-1B \citep{dubey2024llama}, Falcon-7B \citep{almazrouei2023falconseriesopenlanguage}, and Llama3-8B \citep{dubey2024llama}.
We identify that these models share general properties for the next-token probability distributions:
First, as illustrated in \cref{Intro}, the difficulty of approximating a random target distribution is approximately a unimodal function of the target entropy, where the approximation loss increases with the entropy of the target distribution up to a point, then decreases across all models.
Second, for the distributions with moderate entropy, the existence of outlier tokens (i.e., those receiving much higher probability values than the other tokens) makes the approximation easier.
Finally and unsurprisingly, LM-generated distributions are easier to approximate regardless of the entropy values, and can be effectively transferred across LMs.
Our results introduce novel evidence toward a better understanding of the expressiveness of LMs, highlighting the opportunities and challenges in using LMs as probability distribution proposers.

\section{Related Work}
\noindent\textbf{Expressiveness of LMs.}
The output probability distributions of LMs are restricted into a linear subspace of full output space because of softmax bottleneck \citep{yang2018breakingsoftmaxbottleneckhighrank}.
A natural consequence is that any collection of linearly independent LM outputs can form a basis for this space that expresses any probability distribution output by LMs (\citealp{finlayson2023closingcuriouscaseneural}; \citealp{finlayson2024logitsapiprotectedllmsleak}).
While it was previously believed that a large hidden state dimension would overcome the softmax bottleneck in LMs, \citet{chang2022softmax} found that linear dependencies, such as word analogy \citep{Turney_2008}, in token embeddings \citep{mikolov2013distributedrepresentationswordsphrases,ethayarajh-etal-2019-towards} restrict the expressiveness of the output probability distributions.
These findings motivate us to investigate the properties of next token probability distributions to better understand LM expressiveness, which, to the best of our knowledge, has not been systematically studied before.

\begin{figure*}[t]
    \vspace{-10pt}
    \begin{subfigure}[t]{0.33\textwidth}
        \centering
        \includegraphics[width=\textwidth,page=1]{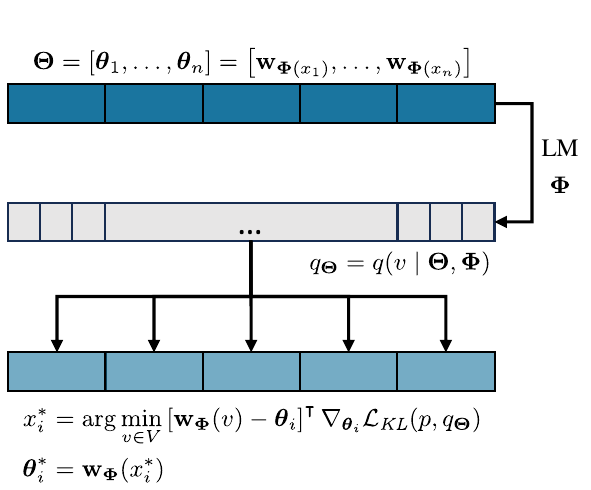}
        \caption{Hard embeddings.}
        \label{fig:hard-framework}
    \end{subfigure}
    \hspace{-20pt}\hfill
    \begin{subfigure}[t]{0.33\textwidth}
        \centering
        \includegraphics[width=\textwidth,page=2]{Framework.pdf}
        \caption{Soft embeddings.}
        \label{fig:soft-framework}
    \end{subfigure}
    \hspace{-20pt}
    \hfill
    \begin{subfigure}[t]{0.33\textwidth}
        \centering
        \includegraphics[width=\textwidth,page=3]{Framework.pdf}
        \caption{Hybrid embeddings.}
        \label{fig:hybrid-framework}
    \end{subfigure}
    \caption{
        \label{fig: frameworks}
        Illustration of the prompt tuning frameworks adapted in this work.
        For hard (\cref{fig:hard-framework}) and hybrid (\cref{fig:hybrid-framework}) embeddings--based frameworks, we optimize the prefix embeddings from left to right; forthe soft embeddings-based framework (\cref{fig:soft-framework}), we optimize the embeddings directly through gradient descent.
        Best viewed in color: the dark colors represent the initial embeddings, and the light colors represent the optimized ones.
    }
    \label{Framework}
\end{figure*}

\vspace{2pt}\noindent\textbf{Prompt tuning with frozen LMs.}
A large body of previous research has demonstrated that prompting can effectively solve a wide range of tasks (\citealp{DBLP:journals/corr/abs-2005-14165}; \citealp{DBLP:journals/corr/abs-2009-07118}; \citealp{DBLP:journals/corr/abs-2012-15723}; \citealp{DBLP:journals/corr/abs-2107-02137}; \citealp{dong2024surveyincontextlearning}, \textit{inter alia}).
Among them, prompt tuning, which learns the prompt parameters from downstream tasks while keeping the main LM parameters frozen, becomes a lightweight yet effective technique to obtain task-specific prompts \citep[cf.][]{DBLP:journals/corr/abs-2107-13586}.
There has been two major branches of tuning methods.
\begin{itemize}[leftmargin=*,topsep=2pt,itemsep=0pt]
    \item \textbf{Hard prompt tuning}, or discrete prompt tuning, which uses the non-contextualized representation of vocabulary tokens as prompts \interalia{ebrahimi2018hotflipwhiteboxadversarialexamples,jiang2020knowlanguagemodelsknow,haviv-etal-2021-bertese}.
          Among them, \citet{wallace2021universaladversarialtriggersattacking} proposed a gradient-based search over token prefixes to trigger certain output tokens.
    \item \textbf{Soft prompt tuning}, or continuous prompt tuning, which finds a sequence of continuous embeddings as the prompt for downstream tasks, while not requiring the prompts to correspond to vocabulary tokens \interalia{li_prefix-tuning_2021,qin2021learningaskqueryinglms,liu2022p}.
          These soft prompts can be directly optimized through gradient descent to the embedding space.
\end{itemize}
Neither branch has been used to study the expressiveness of LMs in eliciting different probability distributions, which is the focus of our work.
In this work, we adapt the approaches of \citet{wallace2021universaladversarialtriggersattacking} and \citet{li_prefix-tuning_2021} to find prompts that lead to specific probability distributions.

\section{Problem Formulation}\label{Problem Statement}
Consider a Transformer-based \citep{DBLP:journals/corr/VaswaniSPUJGKP17} autoregressive language model $\mathcal{M}_{\boldsymbol{\Phi}}$ (e.g., Llama3; \citealp{dubey2024llama}) parametrized by ${\boldsymbol{\Phi}}$.
Let $\mathbf{x} = [x_1, x_2, ..., x_{n}] \in \mathbb{Z}_{\geq 0}^n$ be the input text with $n$ tokens.
The language model $\mathcal{M}_{\boldsymbol{\Phi}}$ first converts $\mathbf{x}$ into non-contextualized token embeddings $\mathbf{W}_{\boldsymbol{\Phi}}(\mathbf{x}) = [\mathbf{w}_{\boldsymbol{\Phi}}(x_1), \ldots, \mathbf{w}_{\boldsymbol{\Phi}}(x_{n})] \in \mathbb{R}^{n \times d}$ through a look-up table, where $d$ is the embedding size.
The embeddings are then fed into the Transformer to produce the logits over the vocabulary $\mathbf{z} = [z_1, \ldots, z_n] \in \mathbb{R}^{|\mathcal{V}|}$, where $|\mathcal{V}|$ is the size of the vocabulary $\mathcal{V}$.
The logits are later on normalized with a softmax operator to produce the probability distribution $\mathbf{q}$ over the vocabulary, where for $v \in \mathcal{V}$, the probability of the next token being $v$ given the prefix $\mathbf{x}$ is given by:\footnote{With a slight abuse of notations, we use $v$ to denote both the token and the token index in the vocabulary.}
\begin{align*}
    q_v = \mathrm{softmax}(\mathbf{z})_v = \frac{\exp(z_v)}{\sum_{v' \in \mathcal{V}} \exp(z_{v'})}.
\end{align*}
Here, $\mathbf{q}$ is essentially the output of the function (i.e., the language model) $\mathcal{M}_{\boldsymbol{\Phi}}$ given the input $\mathbf{x}$; that is, we obtain the next-token distribution $\mathbf{q} \in \mathbb{R}^{|\mathcal{V}|}$ by
\begin{align*}
    \mathbf{q} = \mathcal{M}_{\boldsymbol{\Phi}}(\mathbf{x}).
\end{align*}

We aim to find an input sequence $\mathbf{x}$, where $\mathbf{q} = \mathcal{M}_{\boldsymbol{\Phi}}(\mathbf{x})$ is close enough to an arbitrary target distribution $\mathbf{p} \in \mathbb{R}^{|\mathcal{V}|}$.
Here, since we adopt optimization techniques \citep{li_prefix-tuning_2021,wallace2021universaladversarialtriggersattacking} to find $\mathbf{x}$, $\mathbf{x}$ can be viewed as parameters $\boldsymbol{\Theta}$ ($\boldsymbol{\Theta} = \mathbf{x}$) optimized by the algorithm (\cref{fig: frameworks}).
The objective of the optimization problem is to minimize the KL divergence between the target distribution $\mathbf{p}$ and the output distribution $\mathbf{q}$:
\begin{align*}
                    & \min_{\boldsymbol{\Theta}} \mathcal{L}_{\text{KL}}(\mathbf{p}, \mathbf{q}; {\boldsymbol{\Theta}})                          \\
    \Leftrightarrow & \min_{\boldsymbol{\Theta}} \sum_{v \in \mathcal{V}} p(v) \log \frac{p(v)}{q(v; {\boldsymbol{\Theta}, \boldsymbol{\Phi}})}.
\end{align*}
Lower KL divergence indicates better approximation of the target distributions.
Varying across random seeds, the minimum KL-divergence loss $\mathcal{L}_{\text{KL}}(\mathbf{p}, \mathbf{q}; {\boldsymbol{\Theta}})$ gives an approximation on how difficult it is to elicit the target distribution $\mathbf{p}$ from the LM $\mathcal{M}_{\boldsymbol{\Phi}}$---lower loss indicates easier elicitation.

\section{Methods}
In this section, we introduce our method to synthesize target probability distributions (\cref{sec:target-distribution}) and the prompt tuning frameworks (\cref{sec:prompt-tuning}).

\subsection{Target Distribution Construction}
\label{sec:target-distribution}

Our work requires constructing a probability distribution $\mathbf{p}$ over the vocabulary given a specified entropy value.
To achieve this, we start from a set of random logits $\mathbf{z} \in \mathbb{R}^{|\mathcal{V}|}$, where $\mathcal{V}$ is the vocabulary of the LM.
The probability distribution $p(v)$ defined by the logits $\mathbf{z}$ is given by:
\begin{align*}
    p(v; \mathbf{z}) = \frac{\exp(z_v)}{\sum_{v' \in \mathcal{V}} \exp(z_{v'})}.
\end{align*}
Given the desired target entropy $h$, we aim to find a set of logits $\mathbf{z}$ such that the entropy of the distribution $p(v; \mathbf{z})$ is close to $h$ within a small margin $\varepsilon (\varepsilon > 0)$, by minimizing the following loss with gradient descent:
\begin{align}\label{LossForEntropy}
    \mathcal{L}(\mathbf{z}; h) & = ||H(p(\cdot; \mathbf{z})) - h||^2                                                                                         \\
                               & = \left|\left| -\left(\sum_{v \in \mathcal{V}} p(v; \mathbf{z}) \log p(v; \mathbf{z})\right) - h \right|\right|^2,\nonumber
\end{align}
where $H(\cdot)$ denotes the entropy of a probability distribution.
We use gradient descent to minimize $\mathcal{L}$ to construct for target distribution $\mathbf{p}$.

\begin{figure}[t]
    \includegraphics[width=\columnwidth]{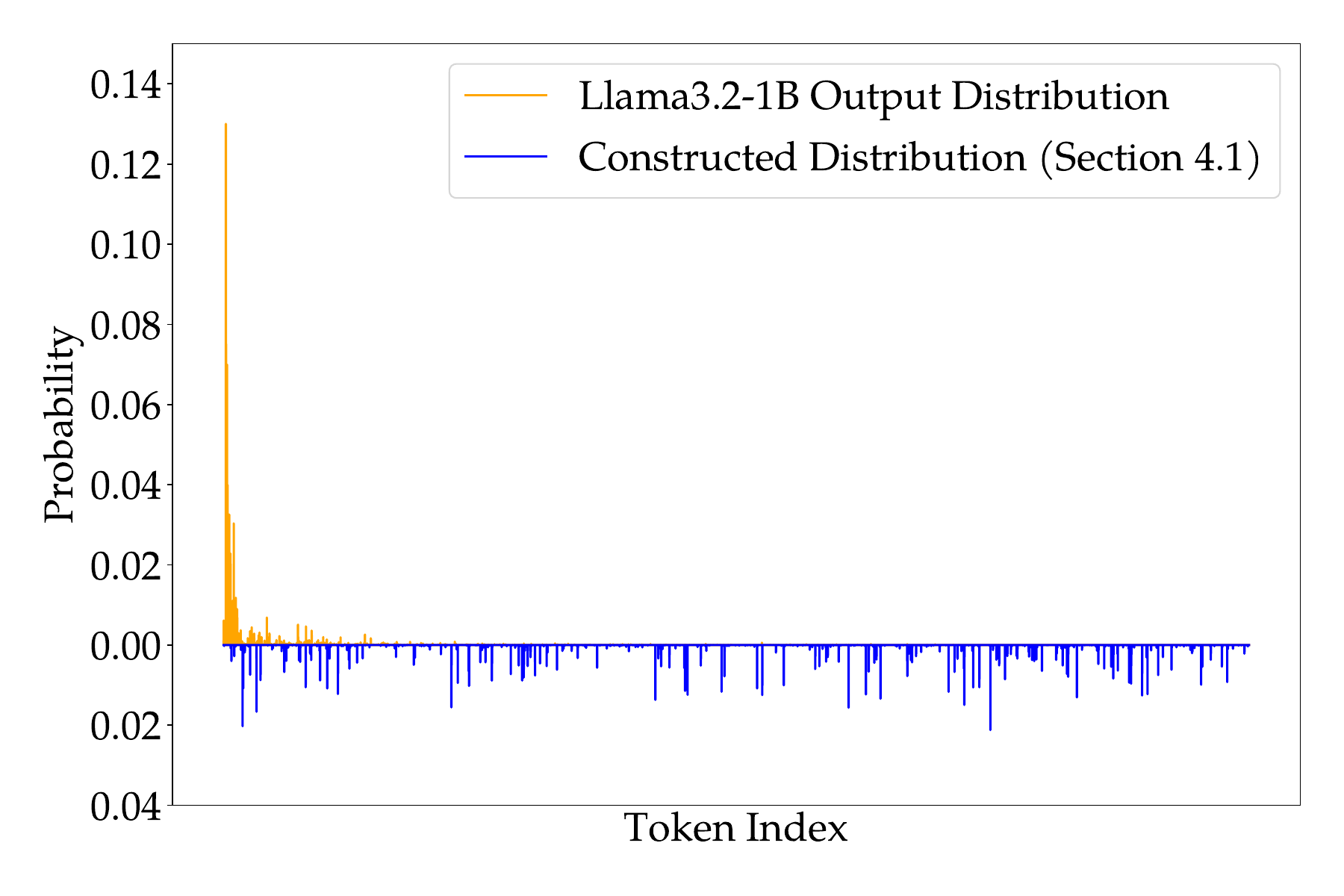}
    \caption{Two distributions, both with entropy 7.68 but have distinctively different shapes. Upper: the output probability distribution from prompting Llama3.2-1B with \textit{'I love'}. Lower: The distribution from our search algorithm following the objective of \cref{LossForEntropy}.}
    \label{Contrast}
\end{figure}

However, entropy is a lossy descriptor of a probability distribution.
For example, in \cref{Contrast}, the two distributions (one output by Llama3.2-1B \citep{dubey2024llama} and the other by our search algorithm based on \cref{LossForEntropy}) have the same entropy yet are distinct from each other sharply.
Considering the fact that LMs usually generate skewed distributions with a few tokens receiving much higher probability mass than the rest \citep{holtzman2020curiouscaseneuraltext}, we define \textit{distributions with outliers}, of which a large amount of probability mass concentrates on a small set of tokens.
Formally, a distribution with $k$ outliers satisfies:
\begin{align}\label{LossForOutlier}
    \sum\limits_{v \in \mathcal{V}_o}p(v) \geq m - \delta,
\end{align}
where $\mathcal{V}_o$ denotes the $k (k\ll|\mathcal{V}|)$ pre-selected outlier tokens that receive distinctively larger probability mass than the rest; $m$ is the theoretical upper bound of the concentrated probability mass to $k$ tokens to ensure the feasibility of the entire distribution reaching the target entropy $h$---details of calculating $m$ from $k$, $h$, and $|\mathcal{V}|$ can be found in \cref{InformationEntropy}; $\delta$ is a small margin to ensure the feasibility of the optimization process, which we set to $0.01$ in all experiments.
Intuitively, \cref{LossForOutlier} requires that a small set of $k$ tokens account for most of the probability mass, thereby capturing our notion of outlier distributions. 
Here, $m$ specifies the maximum feasible cumulative mass these tokens can hold under the entropy constraint, ensuring the construction remains mathematically valid.

We add a regularization term to the loss function in \cref{LossForEntropy} to encourage increased probability mass on the outlier tokens.
To construct an outlier distribution, we optimize the following loss function with gradient descent:
\begin{align}\label{LossForBoth}
    \mathcal{L}'(\mathbf{z}; h) = \alpha \cdot \mathcal{L}(\mathbf{z}; h) - \beta \sum\limits_{v \in \mathcal{V}_o}p(v),
\end{align}
where $\alpha$ and $\beta$ are hyperparameters that balance the two terms, and stop once \cref{LossForOutlier} is satisfied.
We will compare the difficulty of approximating the distributions generated by \cref{LossForEntropy} and \cref{LossForBoth} in our experiments.

\subsection{Prompt Tuning Frameworks}
\label{sec:prompt-tuning}

\noindent\textbf{Hard prompt tuning.}
Intuitively, we aim to find a textual input that elicit the specified target probability distributions.
Following \citet{wallace2021universaladversarialtriggersattacking}, we use an iterative linear approximation of the gradient of the KL divergence loss w.r.t. the prefix embeddings ${\boldsymbol{\Theta}}$.
We update the tokens in the prefix one by one from left to right, where each token $x_i$ is updated to:
\begin{align}
    x_i & = {\arg\min\limits_{x \in \mathcal{V}}} \left[\mathbf{W}_{\boldsymbol{\Phi}}(x) - \boldsymbol{\theta}_i\right]^\intercal \nabla _{\boldsymbol{\theta}_i}\mathcal{L}_{\text{KL}}({\boldsymbol{\Theta}}).
    \label{HardPromptUpdate}
\end{align}
Intuitively, at each step, we find an in-vocabulary token $x \in \mathcal{V}$ that best linearly approximates the gradient of the KL divergence loss w.r.t. the prefix embedding $\boldsymbol{\theta}_i$.
We repeat this iterative update until the convergence of the loss, and obtain the hard prompt tuning result
\begin{align*}
    \boldsymbol{\Theta}^*_\textit{hard} & = \left[\textbf{w}_{\boldsymbol{\Phi}}(x_1), \textbf{w}_{\boldsymbol{\Phi}}(x_2), ..., \textbf{w}_{\boldsymbol{\Phi}}(x_{n-1})\right],
\end{align*}
where $x_i$ is the discrete token that appear in the prefix at position $i$ in the final step.

\vspace{3pt}\noindent\textbf{Soft prompt tuning.}
The power of hard prompts is limited because they are restricted to vocabulary entries.
Following \citet{li_prefix-tuning_2021}, we directly minimize the KL divergence loss w.r.t. the prefix embeddings ${\boldsymbol{\Theta}}$ in a continuous word vector space.
\begin{equation}
    {\boldsymbol{\Theta}}_\textit{soft}^* = \arg \min\limits_{\boldsymbol{\Theta}} \mathcal{L}_{\text{KL}}(\mathbf{p}, \mathbf{q}; {\boldsymbol{\Theta}}).
    \label{SoftPromptUpdate}
\end{equation}
We use AdamW \citep{loshchilov2019decoupledweightdecayregularization} to optimize this objective.
We include soft prompts to probe the upper bound of LM expressivity, since they allow unconstrained optimization in the embedding space; although they may not lie on the natural input manifold, we aim to study theoretical reachability rather than mimic typical usage.

\vspace{3pt}\noindent\textbf{Hybrid prompt tuning.}
To better understand the effect of different tuning methods, we introduce a hybrid tuning method that combines hard and soft prompts, where $\boldsymbol{\Theta}$ can be partitioned into two parts: $\boldsymbol{\Theta}_h$ for hard prompts and $\boldsymbol{\Theta}_s$ for soft prompts, with some $\boldsymbol{\theta}$ updated by hard prompts tuning and others by soft prompts tuning.
We also optimize the hybrid embeddings in a left-to-right manner, where the hard embeddings are updated using \cref{HardPromptUpdate}, with soft embeddings updated using \cref{SoftPromptUpdate}.

We compare the above three strategies in approximating target distributions.
The general framework of our tuning methods can be found in \cref{Framework}.
\section{Experiments}
To study the general property of the expressiveness of an LM, we start by studying different probability distributions of different entropy (\cref{Vanilla Distribution}).
We then analyze the LM approximation performance on outlier distributions (\cref{OutlierExperiment}).
Finally, we investigate target probability distributions as the output distribution of LM itself (\cref{ExactLMOutput}), and analyze the results for approximating the variations of such distributions (\cref{InterveningLMOutput}).
We set up our experiments across transformer-based LMs of different architectures and sizes, including GPT-2 \citep{radford2019language}, Llama3.2-1B \citep{dubey2024llama}, Falcon-7B \citep{almazrouei2023falconseriesopenlanguage}, and Llama3-8B \citep{dubey2024llama}.
To mitigate the effect of randomness brought by target distribution construction algorithms in \cref{sec:target-distribution}, we obtain different distributions regarding the same entropy value.
For each distribution, we use a different initialization for the prefix parameters.
Detailed experiment setups can be found in \cref{ExperimentDetails} for reference.

\subsection{Approximation of Vanilla Distributions}\label{Vanilla Distribution}
We perform prompt tuning to approximate target distributions constructed with \cref{LossForEntropy}, which we will refer to as vanilla distributions.
We investigate the target entropy ranging from 0 to $\log |\mathcal{V}|$ with a step size 0.05, where $|\mathcal{V}|$ is the model's vocabulary size, and $\log |\mathcal{V}|$ is the theoretical upper bound for the vocabulary distribution.\footnote{The proof is detailed in \cref{InformationEntropy}.}

\cref{Intro} presents the approximation loss of soft prompt tuning, plotted against the target entropy for different models.
The approximation loss curve exhibits a consistent pattern across all models: increasing with target entropy up to a point, then decreasing across all models, which roughly forms a unimodal curve.

\begin{figure}[t]
    \centering
    \begin{subfigure}[b]{0.45\textwidth}
        \centering
        \includegraphics[width=\textwidth]{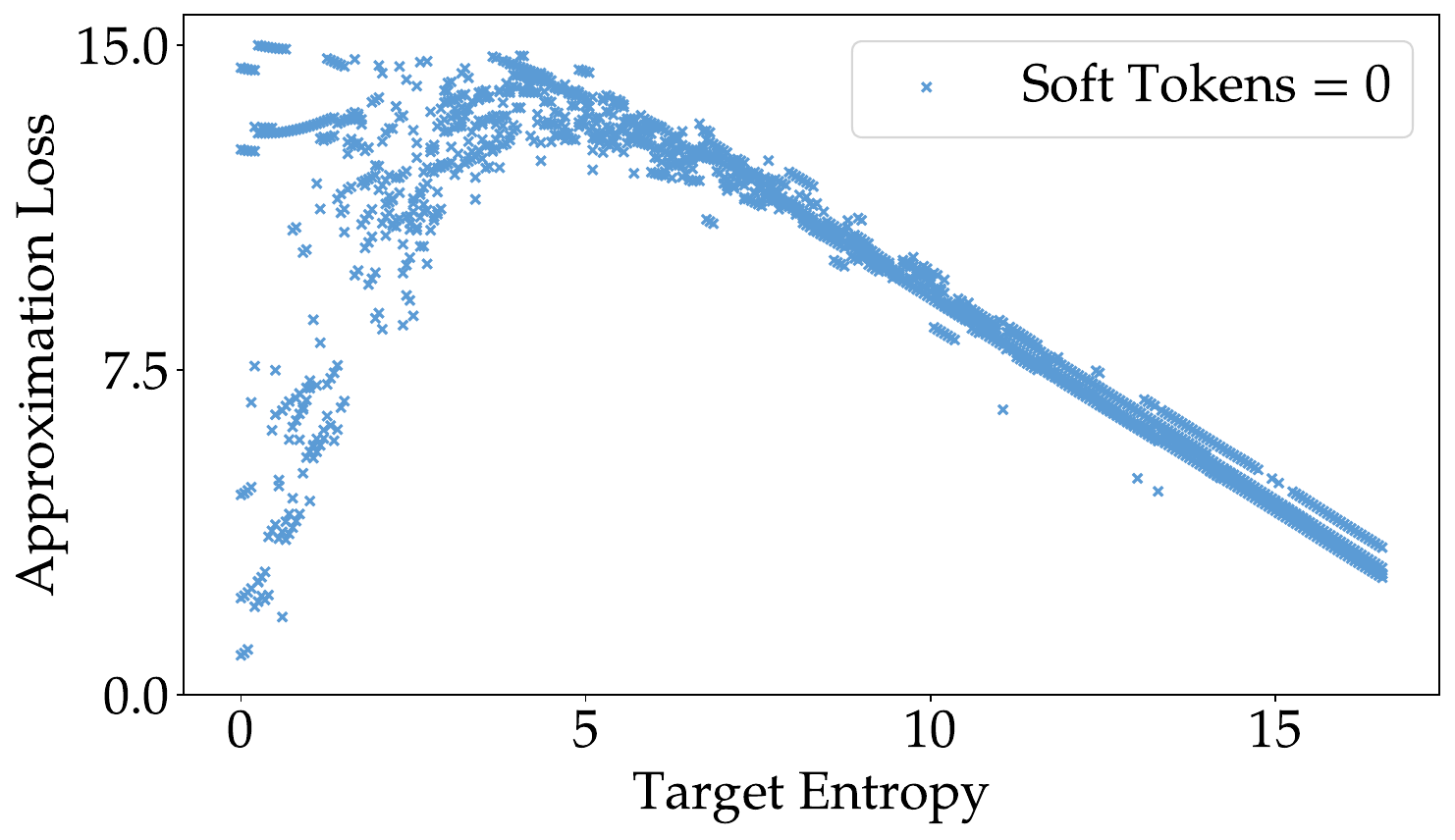}
        \subcaption{Hard prompt tuning.}
        \label{Vanilla_Framework_Compare_Llama3_0}
    \end{subfigure}
    \begin{subfigure}[b]{0.45\textwidth}
        \centering
        \includegraphics[width=\textwidth]{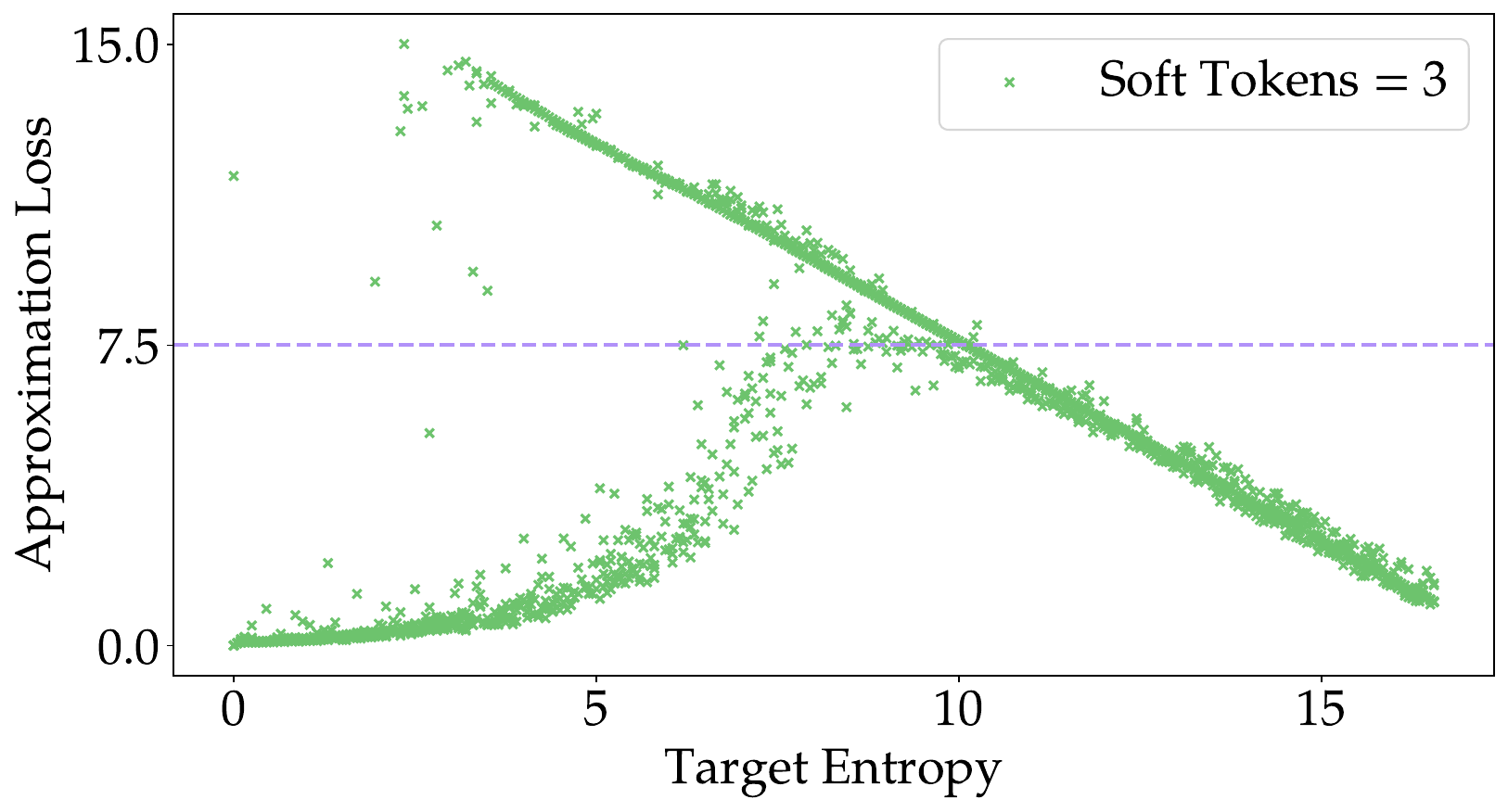}
        \subcaption{Hybrid prompt tuning.}
        \label{Vanilla_Framework_Compare_Llama3_3}
    \end{subfigure}
    \begin{subfigure}[b]{0.45\textwidth}
        \centering
        \includegraphics[width=\textwidth]{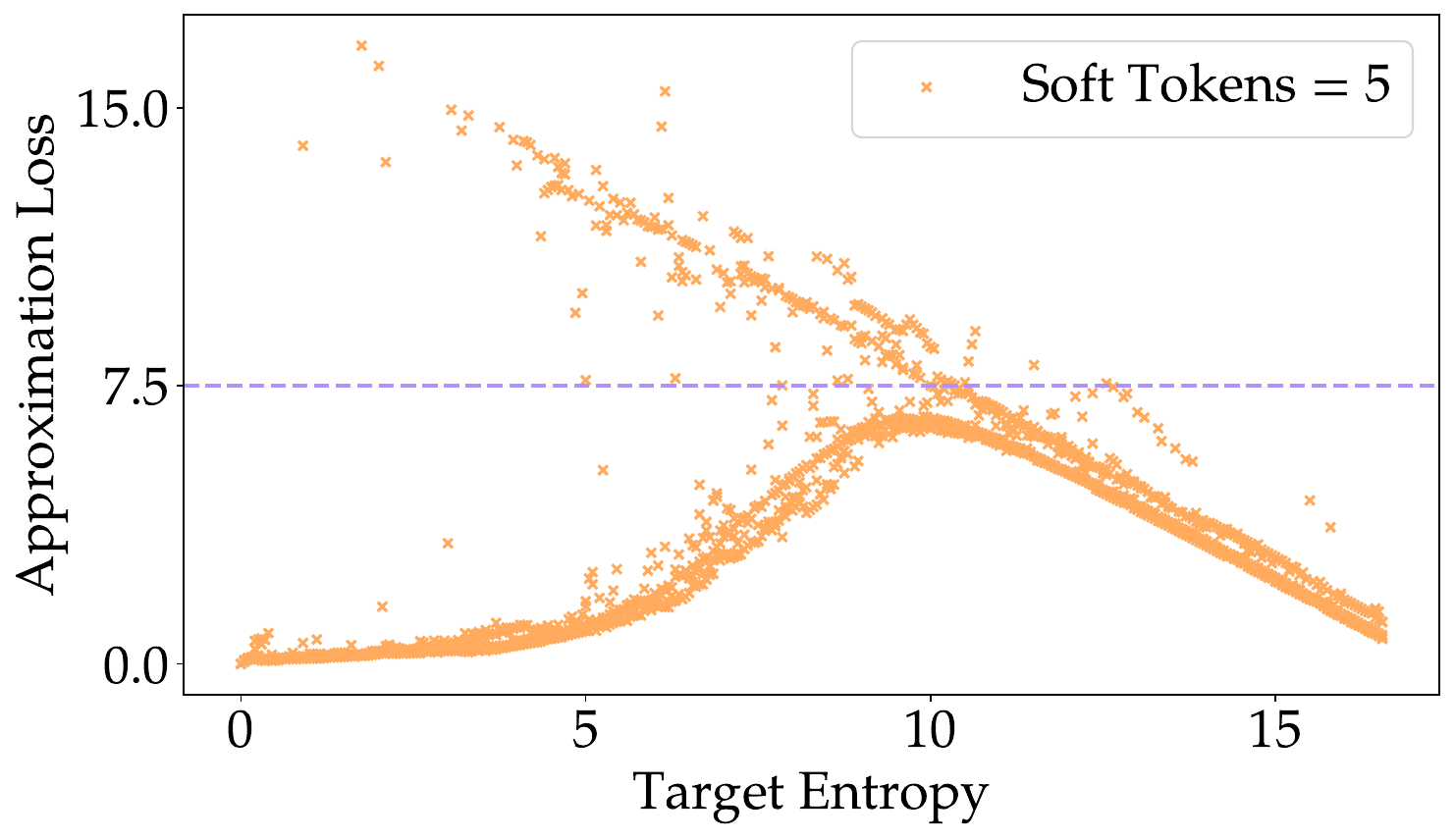}
        \subcaption{Soft prompt tuning.}
        \label{Vanilla_Framework_Compare_Llama3_5}
    \end{subfigure}

    \caption{Approximation loss plotted against the target entropy for different prompt tuning frameworks on Llama3.2-1B. For the distribution of each target entropy we used 5 different prompt initializations, and we keep prefix length to be 5.}
    \label{Vanilla_Framework_Compare_Llama3}
\end{figure}

We then compare soft and hard prompt tuning frameworks (\cref{sec:prompt-tuning}) on vanilla target distributions (\cref{Vanilla_Framework_Compare_Llama3}).
As expected, soft prompts—being unconstrained by the vocabulary—consistently yield lower approximation losses than hard prompts.
To further explore this gap, we adopt hybrid prompt tuning, where updates depend on the embedding source.
Increasing the number of soft tokens in the prefix improves approximation, highlighting the benefit of soft tuning.
We also find that different random initializations of prefix embeddings can lead to large variance in performance.
This effect is especially pronounced in hybrid and soft prompt tuning (\cref{Vanilla_Framework_Compare_Llama3_3,Vanilla_Framework_Compare_Llama3_5}).
Echoing findings in \citet{zhao2025distributional}, our results suggest that randomness plays a systematic role in loss minimization.

\begin{figure}[t]
    \includegraphics[width=\columnwidth]{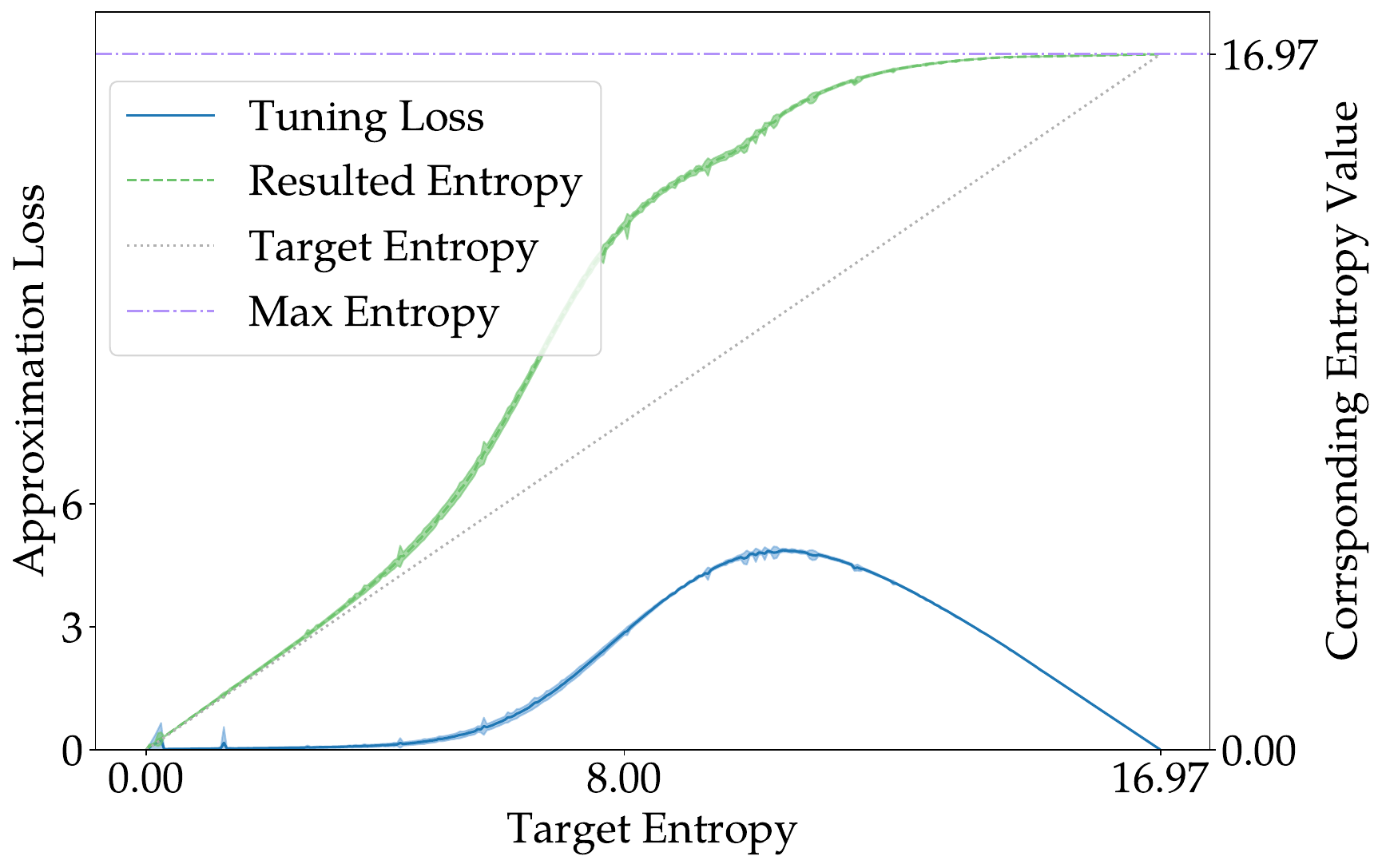}
    \caption{Approximation loss and the entropy for distribution output by tuned prompt plotted against the target entropy for soft prompt tuning on Llama3.2-1B. The figure shows the mean minimum loss within different initializations and corresponding entropy across distributions with the same target entropy (lines) and the 95\% confidence interval (shaded region).
    }
    \label{SoftBaselineLlama3}
\end{figure}

\begin{figure}[t]
    \centering

    \begin{subfigure}[b]{0.48\columnwidth}
        \centering
        \includegraphics[width=\textwidth]{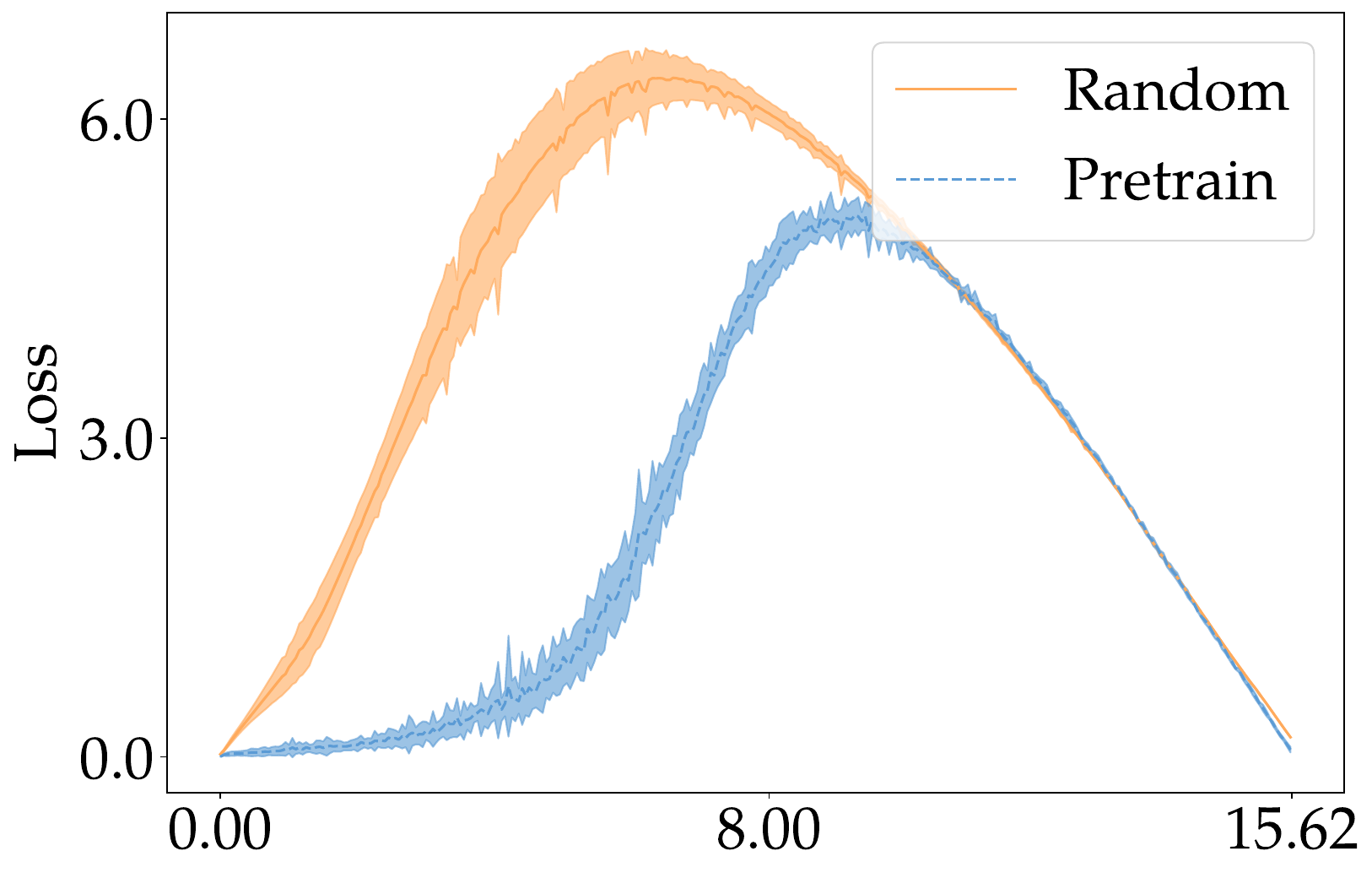}
        \caption{GPT2}
        \label{GPT2}
    \end{subfigure}
    \hfill
    \begin{subfigure}[b]{0.48\columnwidth}
        \centering
        \includegraphics[width=\textwidth]{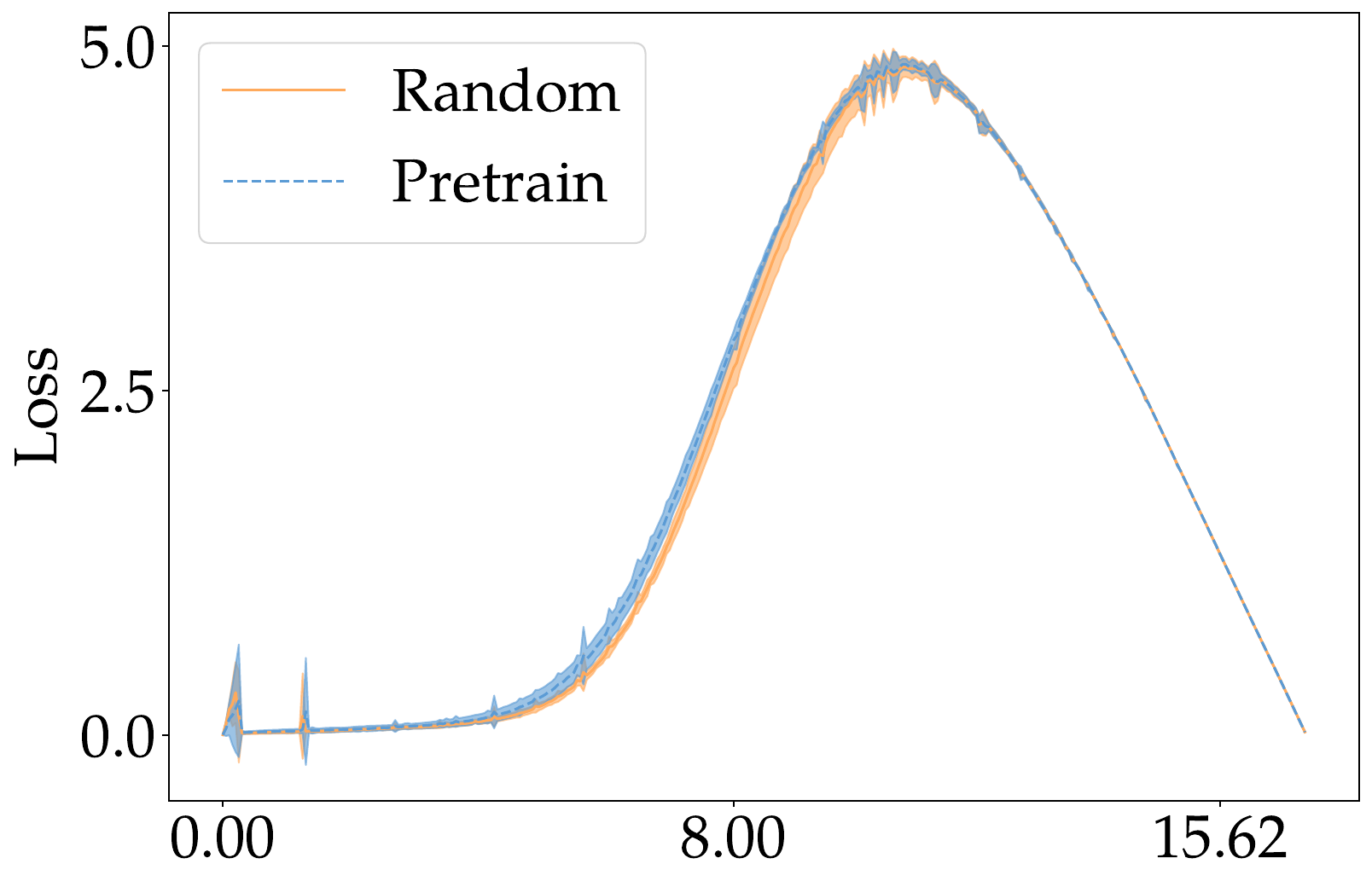}
        \caption{Llama3.2-1B}
        \label{Llama3.2-1B}
    \end{subfigure}

    \vspace{0.5cm}

    \begin{subfigure}[b]{0.48\columnwidth}
        \centering
        \includegraphics[width=\textwidth]{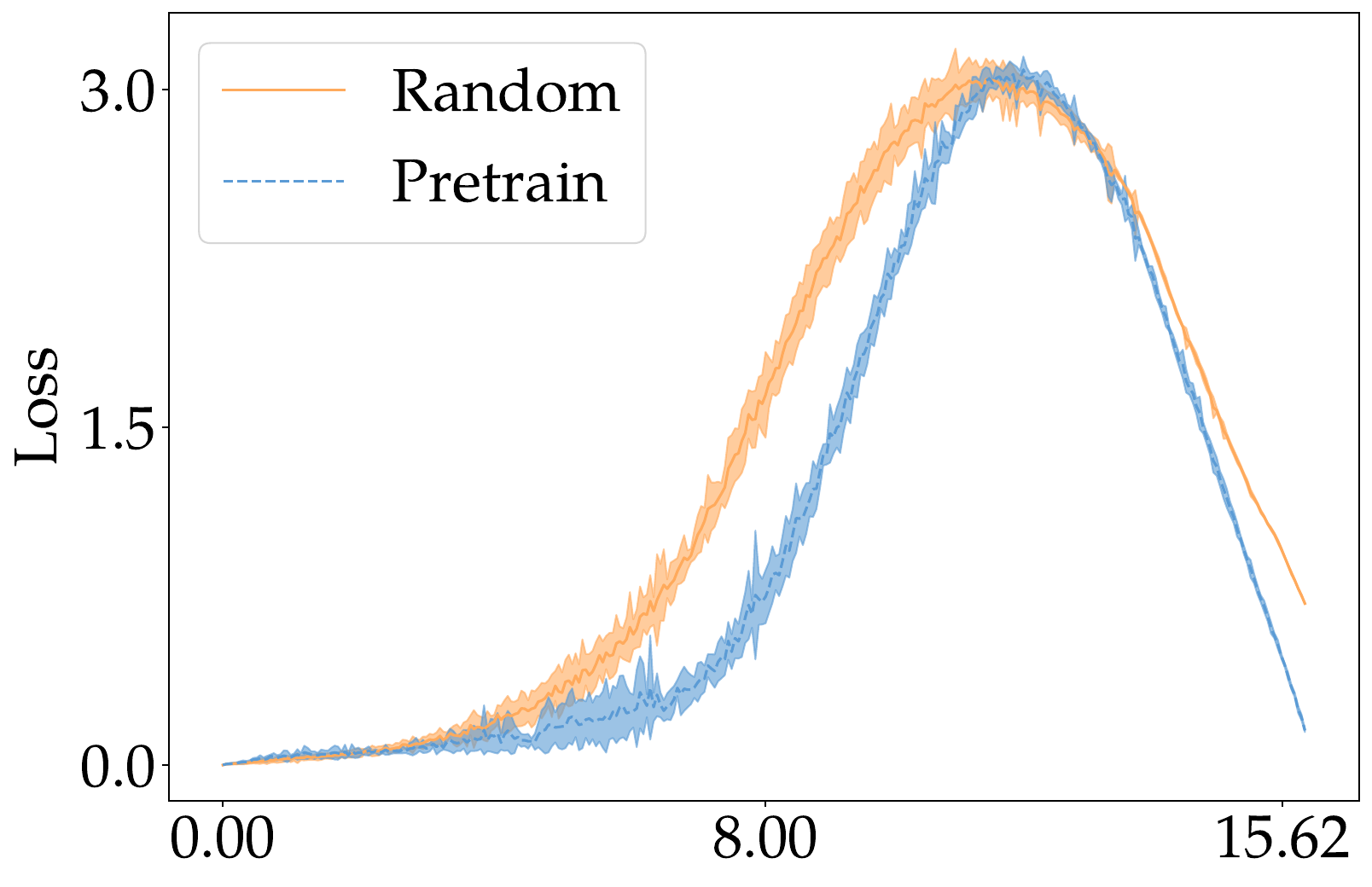}
        \caption{Falcon-7B}
        \label{Falcon-7B}
    \end{subfigure}
    \hfill
    \begin{subfigure}[b]{0.48\columnwidth}
        \centering
        \includegraphics[width=\textwidth]{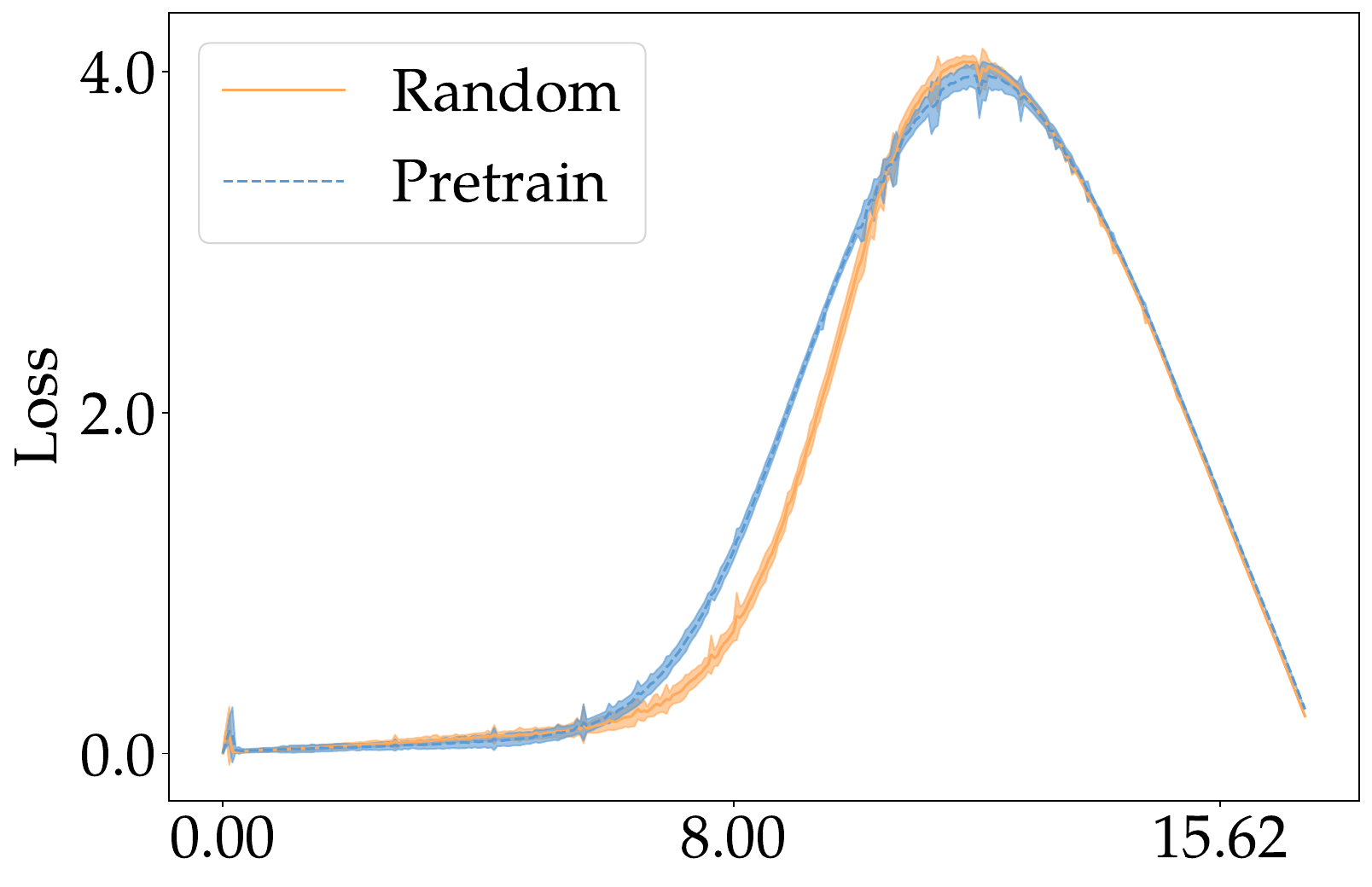}
        \caption{Llama3-8B}
        \label{Llama3-8B}
    \end{subfigure}

    \caption{Approximation losses exhibit a consistent pattern as shown in \cref{Intro} for models randomly initialized that excluded pretraining history. 
    The orange-yellow line is the randomly initiated models, while the blue dash line corresponds to the pretrained models. 
    The x-axis in the subplot stands for target entropy.}
    \label{InitialVanilla}
\end{figure}

However, even when using pure soft prompts for these objectives, we still struggled to elicit certain distributions.
We present the result of a soft prompt tuning study on Llama3.2-1B in \cref{SoftBaselineLlama3}.
The result shows systematic trends of target distributions regarding different ranges of entropy values.
At low entropy, we have near-zero losses on soft prompt approximation, and the prompts tuned by our framework output a distribution with aligned entropy to the target.
However, when the target distribution has a larger entropy, the losses of our training objective increase due to the failure of the model to approximate a lower-entropy distribution from the initializations---the entropy of the approximated distribution grows to be larger than the target entropy.
Such a phenomenon in approximation loss is also observed for transformer-based models from different series and sizes (\cref{Intro}).
When the parameters are randomly initialized, the approximation loss remains a unimodal curve with respect to the target entropy (\cref{InitialVanilla}), indicating that pretraining is not the cause of the unimodal difficulty phenomenon.

\begin{figure}[t]
    \centering
    \begin{subfigure}[b]{0.45\textwidth}
        \centering
        \includegraphics[width=\textwidth]{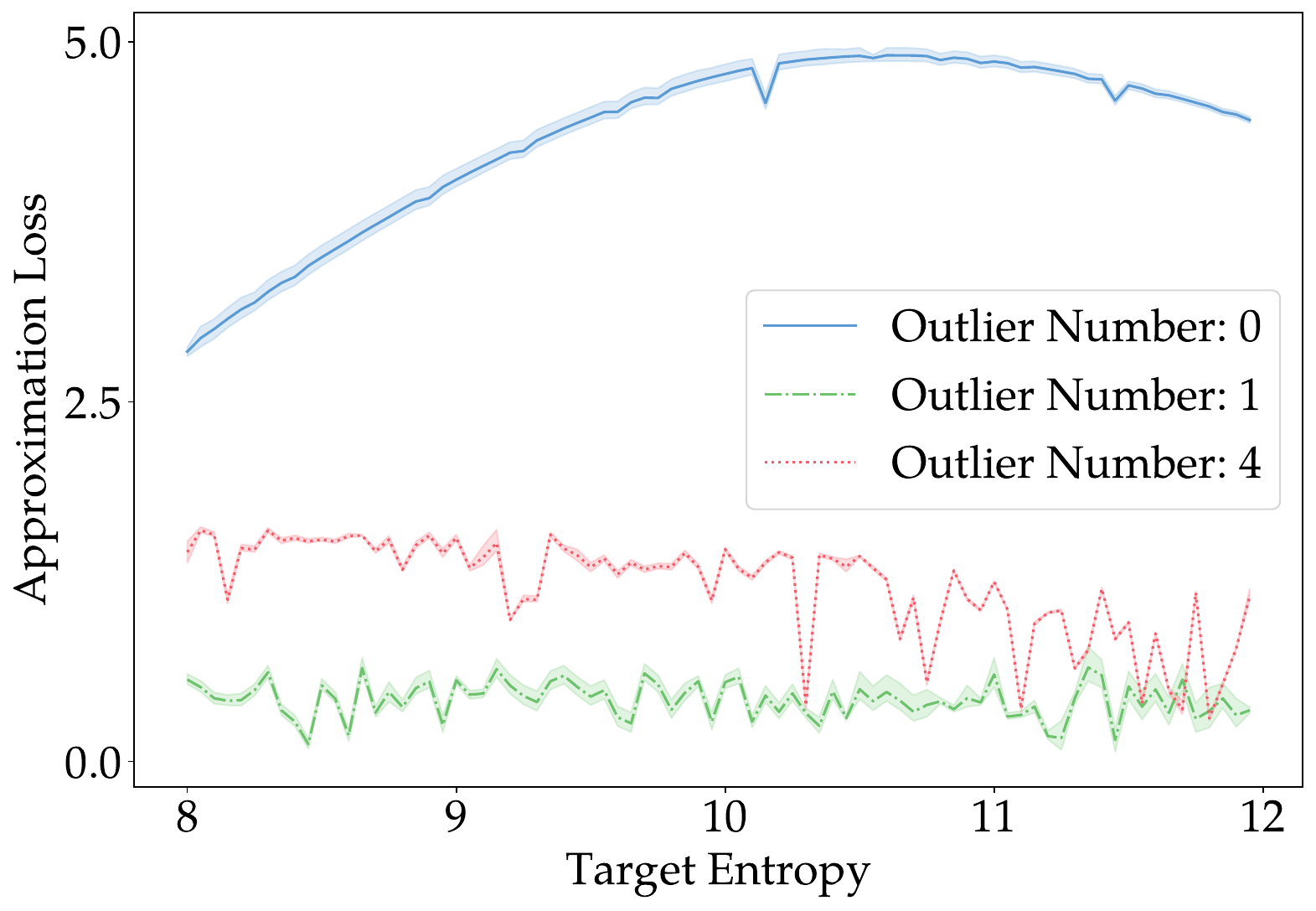}
        \caption{Mean shuffle soft prompt approximation loss on outlier distributions and their 95\% confidence interval (shaded region).}
        \label{OutlierLossLlama3_1B}
    \end{subfigure}
    \hfill
    \begin{subfigure}[b]{0.45\textwidth}
        \centering
        \includegraphics[width=\textwidth]{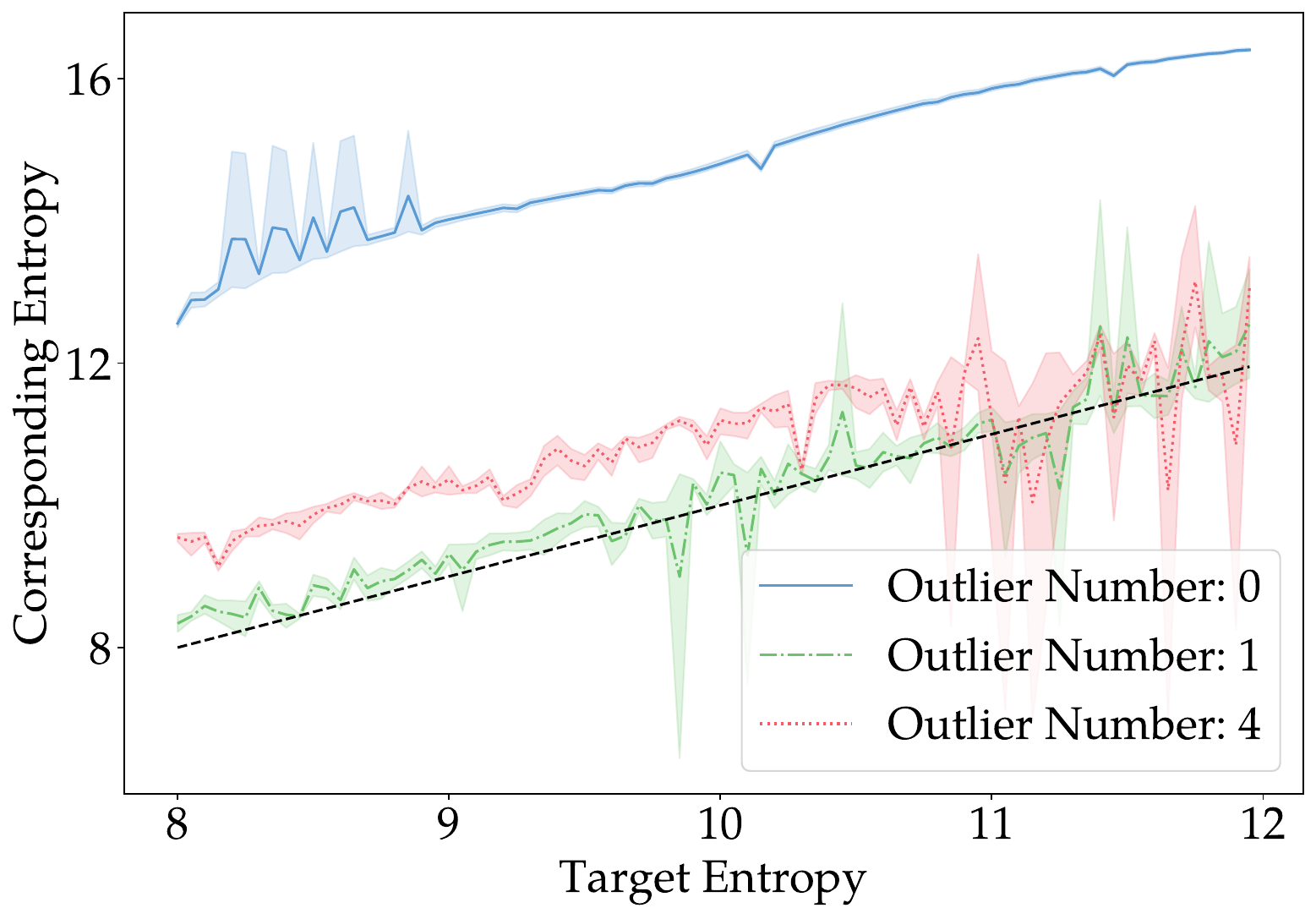}
        \caption{Corresponding mean entropy for shuffle soft prompt approximation on outlier distributions and their 95\% confidence interval (shaded region).}
        \label{OutlierEntropyLlama3_1B}
    \end{subfigure}

    \caption{Outlier distribution loss and corresponding entropy of approximated distributions on Llama3.2-1B.}
    \label{OutlierLoss}
\end{figure}
\begin{figure}[t]
    \centering
    \begin{subfigure}[b]{0.45\textwidth}
        \centering
        \includegraphics[width=\textwidth]{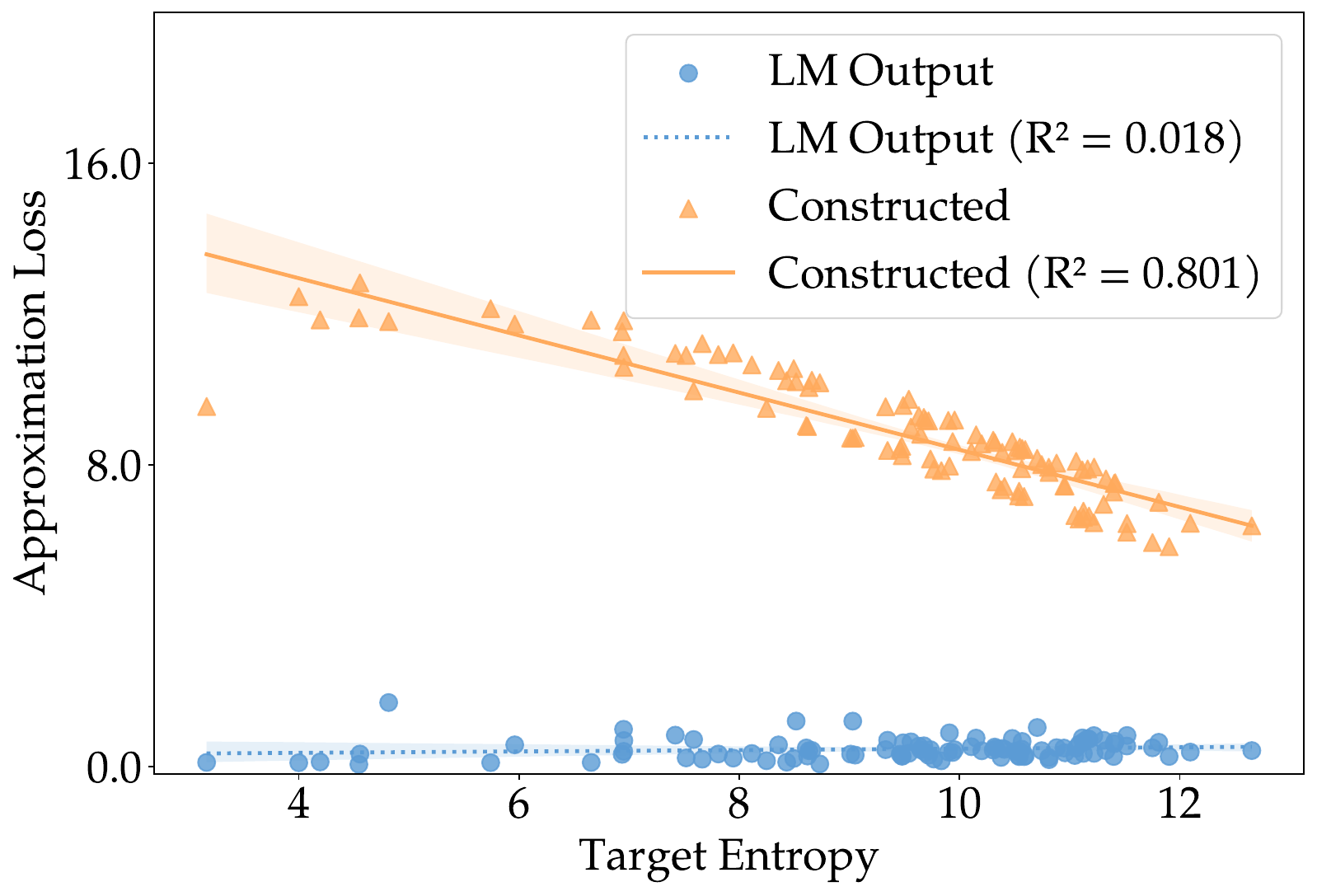}
        \caption{Hard prompt tuning loss comparison between Llama3.2-1B output distributions and matched-entropy vanilla distributions.}
        \label{LMHardLoss}
    \end{subfigure}
    \hfill
    \begin{subfigure}[b]{0.45\textwidth}
        \centering
        \includegraphics[width=\textwidth]{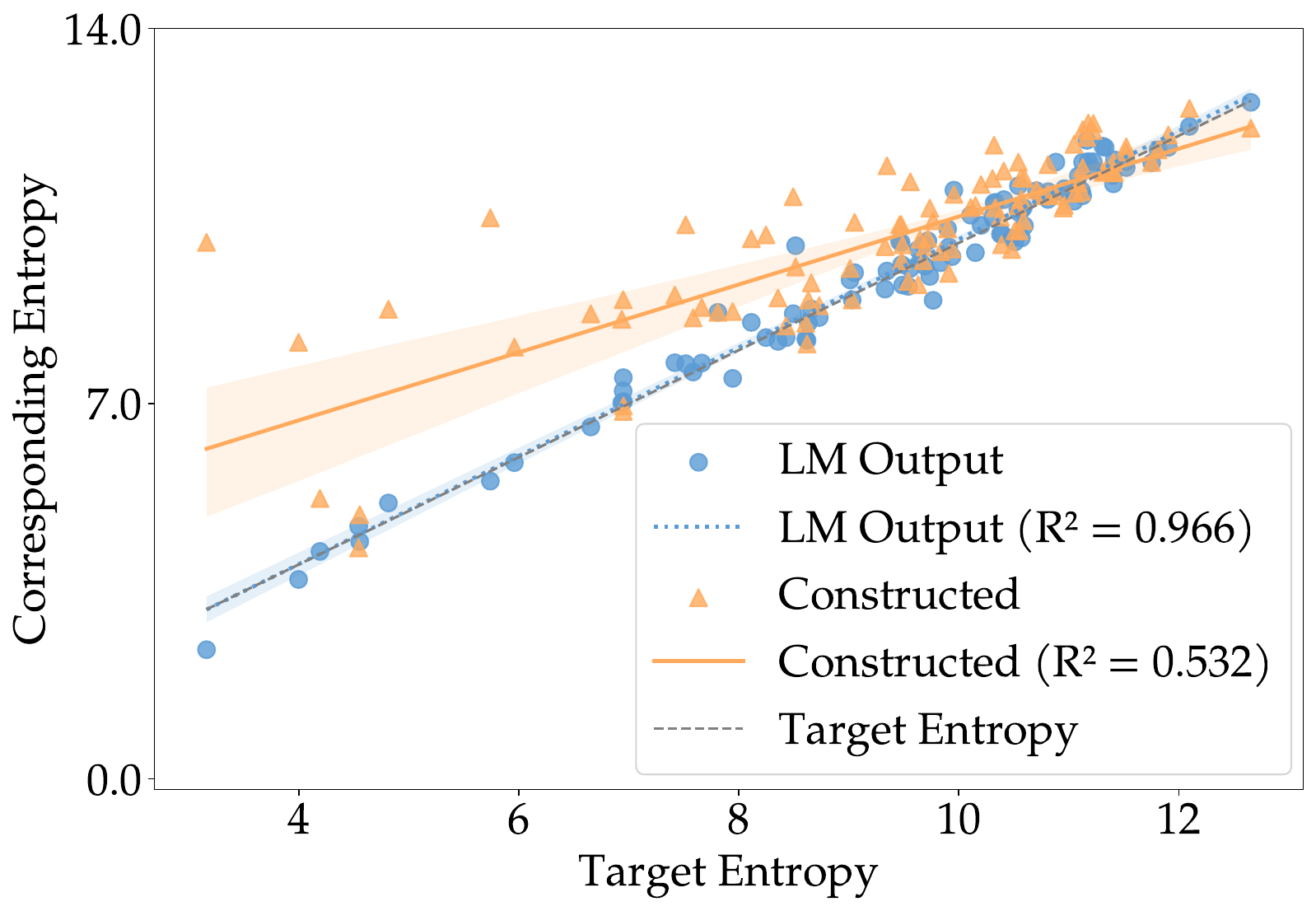}
        \caption{The comparison in corresponding entropy of hard prompt tuning between Llama3.2-1B output distributions and matched-entropy vanilla distributions.}
        \label{LMHardEntropy}
    \end{subfigure}

    \caption{The comparison of approximation performance across distribution pairs in \cref{ExactLMOutput} on Llama3.2-1B.}
    \label{LMTargetLoss}
\end{figure}

\subsection{Approximation of Outlier Distributions} \label{OutlierExperiment}
In this section, we perform prompt tuning to approximate target distributions characterized by outlier tokens, where these distributions are constructed according to the objective defined in \cref{LossForBoth}.

To isolate the effect of outlier tokens from token identity or ordering, we construct five target distributions for each configuration with fixed entropy and outlier count.
To account for variability from prompt initialization, each experiment is repeated with five random initializations, and we report the minimum loss achieved. 
Final results are averaged over these minima across all distributions for each target configuration.

Since moderate-entropy distributions yield higher approximation losses in the vanilla setting (\cref{SoftBaselineLlama3}), we focus our analysis on this regime, where approximation is most challenging.
Our results (\cref{OutlierLossLlama3_1B,OutlierEntropyLlama3_1B}) show that distributions with outlier tokens incur lower losses than vanilla counterparts with similar entropy.
This effect is strongest when a single token dominates the probability mass.
In such cases, the tuned prompts also produce output distributions whose entropy better matches the target.
These findings suggest that the presence of outliers in the target distribution may facilitate more effective prompt tuning by the LM.

\subsection{Approximation of LM-generated Distributions} \label{ExactLMOutput}
Building on the findings of \citet{morris2023languagemodelinversion}, who demonstrate that input text can be largely reconstructed from output distribution logits using a trained inversion model, we hypothesize that LM-generated distributions are intrinsically easier to approximate than distributions we experimented with in \cref{Vanilla Distribution}, irrespective of their entropy levels.

To evaluate this hypothesis, we conduct a comparative analysis between vanilla target distributions (\cref{LossForEntropy}) and target distributions derived from LM outputs.
We conducted our experiments on Llama3.2-1B. 
We construct target distributions by randomly sampling a prompt—i.e., a sequence of random token indices—and using the model's next-token output as the target distribution for evaluation.
We compute the entropy of each LM-generated output distribution and use it as the target entropy in our training objective, as defined in \cref{LossForEntropy}. 
For each, we construct a corresponding vanilla next-token distribution with matched entropy. 
To mitigate initialization variance, we tune 25 prompt initializations per target and report the minimum loss. 
Hard prompt tuning is evaluated on 100 such distributions, each paired with a matched-entropy vanilla distribution.

We present the results in \cref{LMTargetLoss}. 
As shown in \cref{LMHardLoss}, we fit a regression line over the minimum losses and their corresponding entropies across initializations. 
Hard prompt tuning on LM-generated distributions consistently yields lower loss than on vanilla ones, with a low R² value indicating weak correlation between entropy and loss.
\Cref{LMHardEntropy} reports the entropy of the next-token distributions produced by tuned hard prompts for both types of targets.  
The approximated entropy closely aligns with the target for LM-generated distributions, whereas vanilla targets show larger deviations.  

\subsection{Approximation of Variations of LM-Generated Distributions} \label{InterveningLMOutput}
To assess whether LM-generated distributions are not only easier to approximate, as demonstrated in \cref{ExactLMOutput}, but also robust under distributional variations, we extend our analysis to include shuffled LM-generated distributions and distributions derived from the outputs of other models.

\begin{figure}[t]
    \includegraphics[width=\columnwidth]{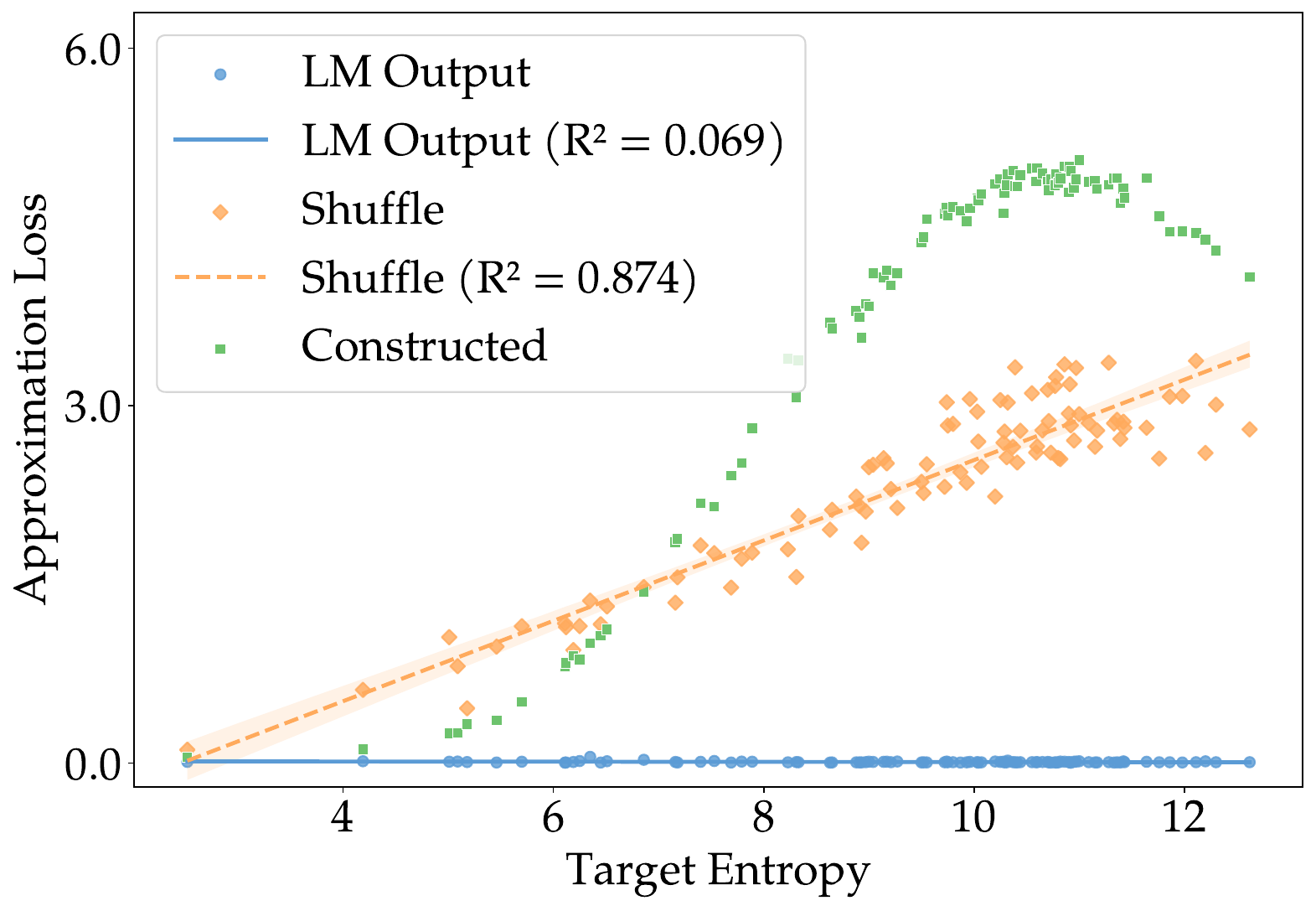}
    \caption{Soft prompt approximation loss is compared across LM-generated, shuffled LM-generated, and vanilla target distributions on Llama3.2-1B.}
    \label{LMShuffleLoss}
\end{figure}

We begin by evaluating how sensitive approximation performance is to token assignments in the target distribution.
To do so, we randomly shuffle token indices in LM-generated distributions from \cref{ExactLMOutput}, preserving their original probabilities.
These are referred to as shuffled variants, with each distribution shuffled five times to reduce randomness.

Following the prior setup, soft prompt tuning is applied using five random initializations, and the minimum loss is reported.
\Cref{LMShuffleLoss} shows the results for shuffled variants alongside original and matched vanilla distributions.
Notably, shuffled variants still outperform vanilla ones—especially in the moderate entropy regime where approximation is harder (\cref{SoftBaselineLlama3}).

\begin{figure}[t]
    \includegraphics[width=\columnwidth]{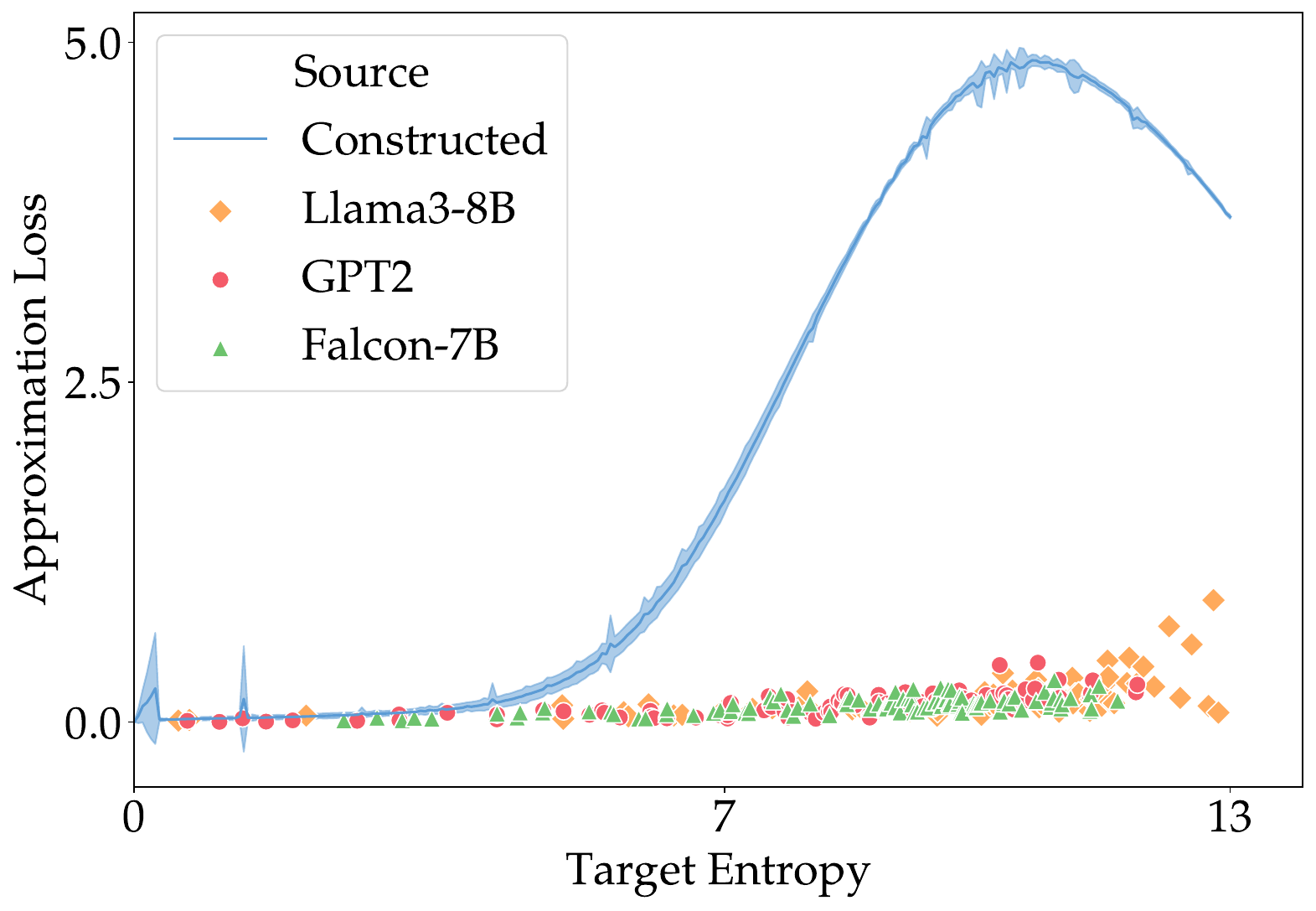}
    \caption{Soft prompt tuning loss on migrated distributions from various source models is evaluated on Llama3.2-1B.}
    \label{MigrationLossLlama3_1B}
\end{figure}

We further examine the transferability of LM-generated distributions by testing whether output probabilities can be effectively approximated across models.
A \textbf{source LM} generates target distributions using the method in \cref{ExactLMOutput}, while a \textbf{tuning LM} performs prompt tuning on these distributions.
The target distribution for the tuning LM is constructed by mapping shared tokens from the source LM and renormalizing their probabilities to ensure a valid distribution.

Soft prompt tuning is used to approximate the migrated distributions from the source models, with Llama3.2-1B as the tuning LM.
Results in \cref{MigrationLossLlama3_1B} show that, despite not being generated by the tuning model, the migrated distributions achieve lower approximation loss than matched-entropy vanilla distributions.
This suggests that, despite differences in tokenization and architecture, LM outputs share underlying properties that support more effective approximation.
\section{Conclusion and Discussion}
In this work, we systematically study the expressiveness of LMs in approximating target next-token probability distributions under prompt tuning frameworks.
We first analyze various prompt tuning methods and find that continuous soft prompts yield the best approximation performance.
From an information-theoretic perspective, we design target distributions with varying entropy and observe that LMs struggle most with those of moderate entropy.
Interestingly, the presence of outlier tokens—rare tokens with notably high probabilities—makes approximation easier.
Moreover, distributions generated by LMs themselves are generally easier to approximate, even under hard prompt tuning, despite having moderate entropy.
These distributions also remain relatively easy to approximate when perturbed.
Overall, our findings reveal general properties of LM expressiveness that can inform more fine-grained future studies.

A natural hypothesis is that the difficulty LMs face in approximating moderate-entropy distributions stems from their training objective: minimizing cross-entropy with a one-hot target encourages low-entropy outputs.
However, our experiments reject this explanation—both randomly initialized and pretrained models exhibit similar behavior.
Moreover, if the hypothesis were correct, approximation loss would increase monotonically with target entropy, which our results do not support.
We hypothesize that such particular difficulty for approximating moderate-entropy distributions may come from either the Transformer \citep{DBLP:journals/corr/VaswaniSPUJGKP17} architecture or the softmax operator in the decoding stage.
We leave the investigation of the root cause of this phenomenon for future work.

While language models are trained to predict the next token, the probabilistic nature of the output distribution makes it naturally a distribution proposer, from the perspectives of subjective randomness \cite{bigelow2023subjective} and computational creativity \cite{peeperkorn2024temperature}, among others. 
Our findings also have implications for interpretability and control: understanding which distributions are harder or easier to elicit can guide robust prompt design, distribution-aware decoding, and fine-grained behavioral analysis (e.g., POS-specific in \Cref{FutureDirections}). 
Beyond prompting, such insights may support failure detection, model evaluation, and the development of decoding strategies that explicitly account for the reachable distribution space. 
Our work provides a first step to systematically analyze and understand the next-token distribution by LMs, and we hope this perspective spurs further exploration of LM expressivity and its applications.

\section*{Limitations}
Our study could be extended in the following ways. 
First, we depend on the investigation of different target probability distributions entirely from the perspective of entropy, which is still a minor step compared to the whole picture of understanding the expressiveness of LMs. 
Second, we notice that the initialization of the prompt tuning actually matters in approximating certain distributions, which is also worthy of research. 
Ultimately, the root cause of the observed phenomenon remains unclear, warranting further investigation to better understand its underlying mechanisms.
\Cref{FutureDirections} presents a case study on instruction-tuned variants and POS-specific distributions.
We hope this work motivates a deeper exploration of LM distributional behavior and guides the design of interpretable models.

\section*{Acknowledgements}
We thank Ziqiao Ma and Jiayuan Mao for constructive discussions at the early stage of this work, as well as the anonymous reviewers and area chairs for the valuable feedback.
This work is supported in part by NSERC RGPIN-2024-04395 and a Canada CIFAR AI Chair award to FS.
\bibliography{custom}

\begin{thebibliography}{35}
\providecommand{\natexlab}[1]{#1}

\bibitem[{Almazrouei et~al.(2023)Almazrouei, Alobeidli, Alshamsi, Cappelli, Cojocaru, Debbah, Étienne Goffinet, Hesslow, Launay, Malartic, Mazzotta, Noune, Pannier, and Penedo}]{almazrouei2023falconseriesopenlanguage}
Ebtesam Almazrouei, Hamza Alobeidli, Abdulaziz Alshamsi, Alessandro Cappelli, Ruxandra Cojocaru, Mérouane Debbah, Étienne Goffinet, Daniel Hesslow, Julien Launay, Quentin Malartic, Daniele Mazzotta, Badreddine Noune, Baptiste Pannier, and Guilherme Penedo. 2023.
\newblock \href {https://arxiv.org/abs/2311.16867} {The falcon series of open language models}.
\newblock \emph{Preprint}, arXiv:2311.16867.

\bibitem[{Bigelow et~al.(2023)Bigelow, Lubana, Dick, Tanaka, and Ullman}]{bigelow2023subjective}
Eric~J Bigelow, Ekdeep~Singh Lubana, Robert~P Dick, Hidenori Tanaka, and Tomer Ullman. 2023.
\newblock Subjective randomness and in-context learning.
\newblock In \emph{UniReps: the First Workshop on Unifying Representations in Neural Models}.

\bibitem[{Bridle(1989)}]{NIPS1989_0336dcba}
John Bridle. 1989.
\newblock \href {https://proceedings.neurips.cc/paper_files/paper/1989/file/0336dcbab05b9d5ad24f4333c7658a0e-Paper.pdf} {Training stochastic model recognition algorithms as networks can lead to maximum mutual information estimation of parameters}.
\newblock In \emph{Advances in Neural Information Processing Systems}, volume~2. Morgan-Kaufmann.

\bibitem[{Brown et~al.(2020)Brown, Mann, Ryder, Subbiah, Kaplan, Dhariwal, Neelakantan, Shyam, Sastry, Askell, Agarwal, Herbert-Voss, Krueger, Henighan, Child, Ramesh, Ziegler, Wu, Winter, Hesse, Chen, Sigler, Litwin, Gray, Chess, Clark, Berner, McCandlish, Radford, Sutskever, and Amodei}]{DBLP:journals/corr/abs-2005-14165}
Tom Brown, Benjamin Mann, Nick Ryder, Melanie Subbiah, Jared~D Kaplan, Prafulla Dhariwal, Arvind Neelakantan, Pranav Shyam, Girish Sastry, Amanda Askell, Sandhini Agarwal, Ariel Herbert-Voss, Gretchen Krueger, Tom Henighan, Rewon Child, Aditya Ramesh, Daniel Ziegler, Jeffrey Wu, Clemens Winter, Chris Hesse, Mark Chen, Eric Sigler, Mateusz Litwin, Scott Gray, Benjamin Chess, Jack Clark, Christopher Berner, Sam McCandlish, Alec Radford, Ilya Sutskever, and Dario Amodei. 2020.
\newblock \href {https://proceedings.neurips.cc/paper_files/paper/2020/file/1457c0d6bfcb4967418bfb8ac142f64a-Paper.pdf} {Language models are few-shot learners}.
\newblock In \emph{Advances in Neural Information Processing Systems}, volume~33, pages 1877--1901. Curran Associates, Inc.

\bibitem[{Chang and McCallum(2022)}]{chang2022softmax}
Haw-Shiuan Chang and Andrew McCallum. 2022.
\newblock \href {https://doi.org/10.18653/v1/2022.acl-long.554} {Softmax bottleneck makes language models unable to represent multi-mode word distributions}.
\newblock In \emph{Proceedings of the 60th Annual Meeting of the Association for Computational Linguistics (Volume 1: Long Papers)}, pages 8048--8073, Dublin, Ireland. Association for Computational Linguistics.

\bibitem[{Dong et~al.(2024)Dong, Li, Dai, Zheng, Ma, Li, Xia, Xu, Wu, Chang, Sun, Li, and Sui}]{dong2024surveyincontextlearning}
Qingxiu Dong, Lei Li, Damai Dai, Ce~Zheng, Jingyuan Ma, Rui Li, Heming Xia, Jingjing Xu, Zhiyong Wu, Baobao Chang, Xu~Sun, Lei Li, and Zhifang Sui. 2024.
\newblock \href {https://doi.org/10.18653/v1/2024.emnlp-main.64} {A survey on in-context learning}.
\newblock In \emph{Proceedings of the 2024 Conference on Empirical Methods in Natural Language Processing}, pages 1107--1128, Miami, Florida, USA. Association for Computational Linguistics.

\bibitem[{Dubey et~al.(2024)Dubey, Jauhri, Pandey, Kadian, Al{-}Dahle, Letman, Mathur, Schelten, Yang, Fan, Goyal, Hartshorn, Yang, Mitra, Sravankumar, Korenev, Hinsvark, Rao, Zhang, Rodriguez, Gregerson, Spataru, Rozi{\`{e}}re, Biron, Tang, Chern, Caucheteux, Nayak, Bi, Marra, McConnell, Keller, Touret, Wu, Wong, Ferrer, Nikolaidis, Allonsius, Song, Pintz, Livshits, Esiobu, Choudhary, Mahajan, Garcia{-}Olano, Perino, Hupkes, Lakomkin, AlBadawy, Lobanova, Dinan, Smith, Radenovic, Zhang, Synnaeve, Lee, Anderson, Nail, Mialon, Pang, Cucurell, Nguyen, Korevaar, Xu, Touvron, Zarov, Ibarra, Kloumann, Misra, Evtimov, Copet, Lee, Geffert, Vranes, Park, Mahadeokar, Shah, van~der Linde, Billock, Hong, Lee, Fu, Chi, Huang, Liu, Wang, Yu, Bitton, Spisak, Park, Rocca, Johnstun, Saxe, Jia, Alwala, Upasani, Plawiak, Li, Heafield, Stone, and et~al.}]{dubey2024llama}
Abhimanyu Dubey, Abhinav Jauhri, Abhinav Pandey, Abhishek Kadian, Ahmad Al{-}Dahle, Aiesha Letman, Akhil Mathur, Alan Schelten, Amy Yang, Angela Fan, Anirudh Goyal, Anthony Hartshorn, Aobo Yang, Archi Mitra, Archie Sravankumar, Artem Korenev, Arthur Hinsvark, Arun Rao, Aston Zhang, Aur{\'{e}}lien Rodriguez, Austen Gregerson, Ava Spataru, Baptiste Rozi{\`{e}}re, Bethany Biron, Binh Tang, Bobbie Chern, Charlotte Caucheteux, Chaya Nayak, Chloe Bi, Chris Marra, Chris McConnell, Christian Keller, Christophe Touret, Chunyang Wu, Corinne Wong, Cristian~Canton Ferrer, Cyrus Nikolaidis, Damien Allonsius, Daniel Song, Danielle Pintz, Danny Livshits, David Esiobu, Dhruv Choudhary, Dhruv Mahajan, Diego Garcia{-}Olano, Diego Perino, Dieuwke Hupkes, Egor Lakomkin, Ehab AlBadawy, Elina Lobanova, Emily Dinan, Eric~Michael Smith, Filip Radenovic, Frank Zhang, Gabriel Synnaeve, Gabrielle Lee, Georgia~Lewis Anderson, Graeme Nail, Gr{\'{e}}goire Mialon, Guan Pang, Guillem Cucurell, Hailey Nguyen, Hannah Korevaar, Hu~Xu, Hugo Touvron, Iliyan Zarov, Imanol~Arrieta Ibarra, Isabel~M. Kloumann, Ishan Misra, Ivan Evtimov, Jade Copet, Jaewon Lee, Jan Geffert, Jana Vranes, Jason Park, Jay Mahadeokar, Jeet Shah, Jelmer van~der Linde, Jennifer Billock, Jenny Hong, Jenya Lee, Jeremy Fu, Jianfeng Chi, Jianyu Huang, Jiawen Liu, Jie Wang, Jiecao Yu, Joanna Bitton, Joe Spisak, Jongsoo Park, Joseph Rocca, Joshua Johnstun, Joshua Saxe, Junteng Jia, Kalyan~Vasuden Alwala, Kartikeya Upasani, Kate Plawiak, Ke~Li, Kenneth Heafield, Kevin Stone, and et~al. 2024.
\newblock \href {https://doi.org/10.48550/ARXIV.2407.21783} {The llama 3 herd of models}.
\newblock \emph{CoRR}, abs/2407.21783.

\bibitem[{Ebrahimi et~al.(2018)Ebrahimi, Rao, Lowd, and Dou}]{ebrahimi2018hotflipwhiteboxadversarialexamples}
Javid Ebrahimi, Anyi Rao, Daniel Lowd, and Dejing Dou. 2018.
\newblock \href {https://doi.org/10.18653/v1/P18-2006} {{H}ot{F}lip: White-box adversarial examples for text classification}.
\newblock In \emph{Proceedings of the 56th Annual Meeting of the Association for Computational Linguistics (Volume 2: Short Papers)}, pages 31--36, Melbourne, Australia. Association for Computational Linguistics.

\bibitem[{Ethayarajh et~al.(2019)Ethayarajh, Duvenaud, and Hirst}]{ethayarajh-etal-2019-towards}
Kawin Ethayarajh, David Duvenaud, and Graeme Hirst. 2019.
\newblock \href {https://doi.org/10.18653/v1/P19-1315} {Towards understanding linear word analogies}.
\newblock In \emph{Proceedings of the 57th Annual Meeting of the Association for Computational Linguistics}, pages 3253--3262, Florence, Italy. Association for Computational Linguistics.

\bibitem[{Finlayson et~al.(2024{\natexlab{a}})Finlayson, Hewitt, Koller, Swayamdipta, and Sabharwal}]{finlayson2023closingcuriouscaseneural}
Matthew Finlayson, John Hewitt, Alexander Koller, Swabha Swayamdipta, and Ashish Sabharwal. 2024{\natexlab{a}}.
\newblock \href {https://openreview.net/forum?id=dONpC9GL1o} {Closing the curious case of neural text degeneration}.
\newblock In \emph{The Twelfth International Conference on Learning Representations, {ICLR} 2024, Vienna, Austria, May 7-11, 2024}. OpenReview.net.

\bibitem[{Finlayson et~al.(2024{\natexlab{b}})Finlayson, Ren, and Swayamdipta}]{finlayson2024logitsapiprotectedllmsleak}
Matthew Finlayson, Xiang Ren, and Swabha Swayamdipta. 2024{\natexlab{b}}.
\newblock \href {https://arxiv.org/abs/2403.09539} {Logits of api-protected llms leak proprietary information}.
\newblock \emph{Preprint}, arXiv:2403.09539.

\bibitem[{Gao et~al.(2021)Gao, Fisch, and Chen}]{DBLP:journals/corr/abs-2012-15723}
Tianyu Gao, Adam Fisch, and Danqi Chen. 2021.
\newblock \href {https://doi.org/10.18653/v1/2021.acl-long.295} {Making pre-trained language models better few-shot learners}.
\newblock In \emph{Proceedings of the 59th Annual Meeting of the Association for Computational Linguistics and the 11th International Joint Conference on Natural Language Processing (Volume 1: Long Papers)}, pages 3816--3830, Online. Association for Computational Linguistics.

\bibitem[{Haviv et~al.(2021)Haviv, Berant, and Globerson}]{haviv-etal-2021-bertese}
Adi Haviv, Jonathan Berant, and Amir Globerson. 2021.
\newblock \href {https://doi.org/10.18653/v1/2021.eacl-main.316} {{BERT}ese: Learning to speak to {BERT}}.
\newblock In \emph{Proceedings of the 16th Conference of the European Chapter of the Association for Computational Linguistics: Main Volume}, pages 3618--3623, Online. Association for Computational Linguistics.

\bibitem[{Holtzman et~al.(2020)Holtzman, Buys, Du, Forbes, and Choi}]{holtzman2020curiouscaseneuraltext}
Ari Holtzman, Jan Buys, Li~Du, Maxwell Forbes, and Yejin Choi. 2020.
\newblock \href {https://openreview.net/forum?id=rygGQyrFvH} {The curious case of neural text degeneration}.
\newblock In \emph{8th International Conference on Learning Representations, {ICLR} 2020, Addis Ababa, Ethiopia, April 26-30, 2020}. OpenReview.net.

\bibitem[{Jiang et~al.(2021)Jiang, Araki, Ding, and Neubig}]{jiang2020knowlanguagemodelsknow}
Zhengbao Jiang, Jun Araki, Haibo Ding, and Graham Neubig. 2021.
\newblock \href {https://doi.org/10.1162/tacl_a_00407} {How can we know when language models know? on the calibration of language models for question answering}.
\newblock \emph{Transactions of the Association for Computational Linguistics}, 9:962--977.

\bibitem[{Kaplan et~al.(2020)Kaplan, McCandlish, Henighan, Brown, Chess, Child, Gray, Radford, Wu, and Amodei}]{kaplan2020scalinglawsneurallanguage}
Jared Kaplan, Sam McCandlish, Tom Henighan, Tom~B. Brown, Benjamin Chess, Rewon Child, Scott Gray, Alec Radford, Jeffrey Wu, and Dario Amodei. 2020.
\newblock \href {https://arxiv.org/abs/2001.08361} {Scaling laws for neural language models}.
\newblock \emph{CoRR}, abs/2001.08361.

\bibitem[{Li and Liang(2021)}]{li_prefix-tuning_2021}
Xiang~Lisa Li and Percy Liang. 2021.
\newblock \href {https://doi.org/10.18653/v1/2021.acl-long.353} {Prefix-tuning: Optimizing continuous prompts for generation}.
\newblock In \emph{Proceedings of the 59th Annual Meeting of the Association for Computational Linguistics and the 11th International Joint Conference on Natural Language Processing (Volume 1: Long Papers)}, pages 4582--4597, Online. Association for Computational Linguistics.

\bibitem[{Liu et~al.(2023)Liu, Yuan, Fu, Jiang, Hayashi, and Neubig}]{DBLP:journals/corr/abs-2107-13586}
Pengfei Liu, Weizhe Yuan, Jinlan Fu, Zhengbao Jiang, Hiroaki Hayashi, and Graham Neubig. 2023.
\newblock \href {https://doi.org/10.1145/3560815} {Pre-train, prompt, and predict: {A} systematic survey of prompting methods in natural language processing}.
\newblock \emph{{ACM} Comput. Surv.}, 55(9):195:1--195:35.

\bibitem[{Liu et~al.(2022)Liu, Ji, Fu, Tam, Du, Yang, and Tang}]{liu2022p}
Xiao Liu, Kaixuan Ji, Yicheng Fu, Weng Tam, Zhengxiao Du, Zhilin Yang, and Jie Tang. 2022.
\newblock P-tuning: Prompt tuning can be comparable to fine-tuning across scales and tasks.
\newblock In \emph{Proceedings of the 60th Annual Meeting of the Association for Computational Linguistics (Volume 2: Short Papers)}, pages 61--68.

\bibitem[{Loshchilov and Hutter(2019)}]{loshchilov2019decoupledweightdecayregularization}
Ilya Loshchilov and Frank Hutter. 2019.
\newblock \href {https://openreview.net/forum?id=Bkg6RiCqY7} {Decoupled weight decay regularization}.
\newblock In \emph{7th International Conference on Learning Representations, {ICLR} 2019, New Orleans, LA, USA, May 6-9, 2019}. OpenReview.net.

\bibitem[{Mikolov et~al.(2013)Mikolov, Sutskever, Chen, Corrado, and Dean}]{mikolov2013distributedrepresentationswordsphrases}
Tom{\'{a}}s Mikolov, Ilya Sutskever, Kai Chen, Gregory~S. Corrado, and Jeffrey Dean. 2013.
\newblock \href {https://proceedings.neurips.cc/paper/2013/hash/9aa42b31882ec039965f3c4923ce901b-Abstract.html} {Distributed representations of words and phrases and their compositionality}.
\newblock In \emph{Advances in Neural Information Processing Systems 26: 27th Annual Conference on Neural Information Processing Systems 2013. Proceedings of a meeting held December 5-8, 2013, Lake Tahoe, Nevada, United States}, pages 3111--3119.

\bibitem[{Morris et~al.(2024)Morris, Zhao, Chiu, Shmatikov, and Rush}]{morris2023languagemodelinversion}
John~X. Morris, Wenting Zhao, Justin~T. Chiu, Vitaly Shmatikov, and Alexander~M. Rush. 2024.
\newblock \href {https://openreview.net/forum?id=t9dWHpGkPj} {Language model inversion}.
\newblock In \emph{The Twelfth International Conference on Learning Representations, {ICLR} 2024, Vienna, Austria, May 7-11, 2024}. OpenReview.net.

\bibitem[{Neumann et~al.(2019)Neumann, King, Beltagy, and Ammar}]{DBLP:conf/bionlp/NeumannKBA19}
Mark Neumann, Daniel King, Iz~Beltagy, and Waleed Ammar. 2019.
\newblock \href {https://doi.org/10.18653/V1/W19-5034} {Scispacy: Fast and robust models for biomedical natural language processing}.
\newblock In \emph{Proceedings of the 18th BioNLP Workshop and Shared Task, BioNLP@ACL 2019, Florence, Italy, August 1, 2019}, pages 319--327. Association for Computational Linguistics.

\bibitem[{Peeperkorn et~al.(2024)Peeperkorn, Kouwenhoven, Brown, and Jordanous}]{peeperkorn2024temperature}
Max Peeperkorn, Tom Kouwenhoven, Dan Brown, and Anna Jordanous. 2024.
\newblock Is temperature the creativity parameter of large language models?
\newblock \emph{arXiv preprint arXiv:2405.00492}.

\bibitem[{Qin and Eisner(2021)}]{qin2021learningaskqueryinglms}
Guanghui Qin and Jason Eisner. 2021.
\newblock \href {https://doi.org/10.18653/v1/2021.naacl-main.410} {Learning how to ask: Querying {LM}s with mixtures of soft prompts}.
\newblock In \emph{Proceedings of the 2021 Conference of the North American Chapter of the Association for Computational Linguistics: Human Language Technologies}, pages 5203--5212, Online. Association for Computational Linguistics.

\bibitem[{Radford et~al.(2019)Radford, Wu, Child, Luan, Amodei, and Sutskever}]{radford2019language}
Alec Radford, Jeffrey Wu, Rewon Child, David Luan, Dario Amodei, and Ilya Sutskever. 2019.
\newblock \href {https://cdn.openai.com/better-language-models/language_models_are_unsupervised_multitask_learners.pdf} {Language models are unsupervised multitask learners}.
\newblock \emph{OpenAI}.

\bibitem[{Schick and Sch{\"{u}}tze(2021)}]{DBLP:journals/corr/abs-2009-07118}
Timo Schick and Hinrich Sch{\"{u}}tze. 2021.
\newblock \href {https://doi.org/10.18653/V1/2021.NAACL-MAIN.185} {It's not just size that matters: Small language models are also few-shot learners}.
\newblock In \emph{Proceedings of the 2021 Conference of the North American Chapter of the Association for Computational Linguistics: Human Language Technologies, {NAACL-HLT} 2021, Online, June 6-11, 2021}, pages 2339--2352. Association for Computational Linguistics.

\bibitem[{Sun et~al.(2021)Sun, Wang, Feng, Ding, Pang, Shang, Liu, Chen, Zhao, Lu, Liu, Wu, Gong, Liang, Shang, Sun, Liu, Ouyang, Yu, Tian, Wu, and Wang}]{DBLP:journals/corr/abs-2107-02137}
Yu~Sun, Shuohuan Wang, Shikun Feng, Siyu Ding, Chao Pang, Junyuan Shang, Jiaxiang Liu, Xuyi Chen, Yanbin Zhao, Yuxiang Lu, Weixin Liu, Zhihua Wu, Weibao Gong, Jianzhong Liang, Zhizhou Shang, Peng Sun, Wei Liu, Xuan Ouyang, Dianhai Yu, Hao Tian, Hua Wu, and Haifeng Wang. 2021.
\newblock \href {https://arxiv.org/abs/2107.02137} {{ERNIE} 3.0: Large-scale knowledge enhanced pre-training for language understanding and generation}.
\newblock \emph{CoRR}, abs/2107.02137.

\bibitem[{Turney(2008)}]{Turney_2008}
P.~D. Turney. 2008.
\newblock \href {https://doi.org/10.1613/jair.2693} {The latent relation mapping engine: Algorithm and experiments}.
\newblock \emph{Journal of Artificial Intelligence Research}, 33:615–655.

\bibitem[{Vaswani et~al.(2017)Vaswani, Shazeer, Parmar, Uszkoreit, Jones, Gomez, Kaiser, and Polosukhin}]{DBLP:journals/corr/VaswaniSPUJGKP17}
Ashish Vaswani, Noam Shazeer, Niki Parmar, Jakob Uszkoreit, Llion Jones, Aidan~N. Gomez, Lukasz Kaiser, and Illia Polosukhin. 2017.
\newblock \href {https://proceedings.neurips.cc/paper/2017/hash/3f5ee243547dee91fbd053c1c4a845aa-Abstract.html} {Attention is all you need}.
\newblock In \emph{Advances in Neural Information Processing Systems 30: Annual Conference on Neural Information Processing Systems 2017, December 4-9, 2017, Long Beach, CA, {USA}}, pages 5998--6008.

\bibitem[{Wallace et~al.(2019)Wallace, Feng, Kandpal, Gardner, and Singh}]{wallace2021universaladversarialtriggersattacking}
Eric Wallace, Shi Feng, Nikhil Kandpal, Matt Gardner, and Sameer Singh. 2019.
\newblock \href {https://doi.org/10.18653/v1/D19-1221} {Universal adversarial triggers for attacking and analyzing {NLP}}.
\newblock In \emph{Proceedings of the 2019 Conference on Empirical Methods in Natural Language Processing and the 9th International Joint Conference on Natural Language Processing (EMNLP-IJCNLP)}, pages 2153--2162, Hong Kong, China. Association for Computational Linguistics.

\bibitem[{Yang et~al.(2024)Yang, Yang, Zhang, Hui, Zheng, Yu, Li, Liu, Huang, Wei, Lin, Yang, Tu, Zhang, Yang, Yang, Zhou, Lin, Dang, Lu, Bao, Yang, Yu, Li, Xue, Zhang, Zhu, Men, Lin, Li, Xia, Ren, Ren, Fan, Su, Zhang, Wan, Liu, Cui, Zhang, and Qiu}]{DBLP:journals/corr/abs-2412-15115}
An~Yang, Baosong Yang, Beichen Zhang, Binyuan Hui, Bo~Zheng, Bowen Yu, Chengyuan Li, Dayiheng Liu, Fei Huang, Haoran Wei, Huan Lin, Jian Yang, Jianhong Tu, Jianwei Zhang, Jianxin Yang, Jiaxi Yang, Jingren Zhou, Junyang Lin, Kai Dang, Keming Lu, Keqin Bao, Kexin Yang, Le~Yu, Mei Li, Mingfeng Xue, Pei Zhang, Qin Zhu, Rui Men, Runji Lin, Tianhao Li, Tingyu Xia, Xingzhang Ren, Xuancheng Ren, Yang Fan, Yang Su, Yichang Zhang, Yu~Wan, Yuqiong Liu, Zeyu Cui, Zhenru Zhang, and Zihan Qiu. 2024.
\newblock \href {https://doi.org/10.48550/ARXIV.2412.15115} {Qwen2.5 technical report}.
\newblock \emph{CoRR}, abs/2412.15115.

\bibitem[{Yang et~al.(2018)Yang, Dai, Salakhutdinov, and Cohen}]{yang2018breakingsoftmaxbottleneckhighrank}
Zhilin Yang, Zihang Dai, Ruslan Salakhutdinov, and William~W. Cohen. 2018.
\newblock \href {https://openreview.net/forum?id=HkwZSG-CZ} {Breaking the softmax bottleneck: {A} high-rank {RNN} language model}.
\newblock In \emph{6th International Conference on Learning Representations, {ICLR} 2018, Vancouver, BC, Canada, April 30 - May 3, 2018, Conference Track Proceedings}. OpenReview.net.

\bibitem[{Zhao et~al.(2025)Zhao, Qin, Alvarez-Melis, Kakade, and Saphra}]{zhao2025distributional}
Rosie Zhao, Tian Qin, David Alvarez-Melis, Sham Kakade, and Naomi Saphra. 2025.
\newblock Distributional scaling laws for emergent capabilities.
\newblock \emph{arXiv preprint arXiv:2502.17356}.

\bibitem[{Zhao et~al.(2023)Zhao, Zhou, Li, Tang, Wang, Hou, Min, Zhang, Zhang, Dong, Du, Yang, Chen, Chen, Jiang, Ren, Li, Tang, Liu, Liu, Nie, and Wen}]{zhao2024surveylargelanguagemodels}
Wayne~Xin Zhao, Kun Zhou, Junyi Li, Tianyi Tang, Xiaolei Wang, Yupeng Hou, Yingqian Min, Beichen Zhang, Junjie Zhang, Zican Dong, Yifan Du, Chen Yang, Yushuo Chen, Zhipeng Chen, Jinhao Jiang, Ruiyang Ren, Yifan Li, Xinyu Tang, Zikang Liu, Peiyu Liu, Jian{-}Yun Nie, and Ji{-}Rong Wen. 2023.
\newblock \href {https://doi.org/10.48550/ARXIV.2303.18223} {A survey of large language models}.
\newblock \emph{CoRR}, abs/2303.18223.

\end{thebibliography}
\bibliographystyle{acl_natbib}
\newpage
\appendix
\section{Entropy} \label{InformationEntropy}
For a probability distribution $p = (p_1, p_2, ..., p_n)$, the entropy for this probability distribution is defined as:

$$
    H(p) = -\sum\limits_{i=1}^np_i \log(p_i)
$$
while $p_i$ represents the probability value of event $i$ and $\sum\limits_{i=1}^np_i = 1$.

For given constraints that the sum of the probability value for each event is 1, the probability distribution $p$ that maximizes entropy is the uniform distribution. To prove this, we introduce a Lagrange multiplier $\lambda$ for this constraint and form the Lagrange function:

$$
    \mathcal{L}(p_1, p_2, ..., p_n, \lambda) = -\sum\limits_{i=1}^np_i\log p_i + \lambda(\sum\limits_{i=1}^np_i - 1)
$$

To maximize $\mathcal{L}$, we take partial derivative of $\mathcal{L}$ with respect to each $p_i$ and set it equal to 0:

$$
    \frac{\partial \mathcal{L}}{\partial p_i} = -\log p_i - 1 + \lambda = 0.
$$

Since this must hold true for each $p_i$, we conclude that each $p_i$ should be the same. To be more specific, we can conclude that to take the maximum value, $p_i$ must equal to $c$, where $c$ is a constant. To make this satisfy the constraints previously, we have:

$$
    p_i = \frac{1}{n}.
$$

Therefore, the probability distribution that maximizes entropy is the uniform distribution, where each $p_i = \frac{1}{n}$, and the maximum entropy for a distribution is $\log(n)$.

We now establish the detailed calculation for value $m$ defined as the upper boundary for probability mass concentrate on outlier tokens in \cref{LossForBoth}. Because of the linearity of entropy $H(p)$, we take the probability distribution of outlier tokens and the rest into separate consideration. For $k$ outlier tokens, the maximum entropy is reached when there is uniform distribution for them. As we assume that $m$ probability mass has fallen on these tokens, the maximum entropy for outlier tokens is $-m \log(\frac{m}{k})$. Similarly, for the rest tokens, the maximum entropy is $-(1-m)\log (\frac{1-m}{n-k})$. So in case we desire to find a probability distribution with the existence of $k$ outliers to meet a specific entropy $e$, we must satisfy the inequality:

$$
    -m \log(\frac{m}{k}) - (1-m)\log (\frac{1-m}{n-k}) \geq e.
$$

Also according to the definition of outliers, we notice that $k \ll n$. So given the outlier number $k$ fixed, the LHS of the inequality decreases as outlier probability $m$ increases. Also noticed that for a set of small enough $k$, the most restrictive condition is achieved when $k = 1$. Then, we change our objective to solve the equation as:

$$
    -m \log(m) - (1-m)\log (\frac{1-m}{n-1}) = e.
$$

\begin{figure}[t]
    \includegraphics[width=\columnwidth]{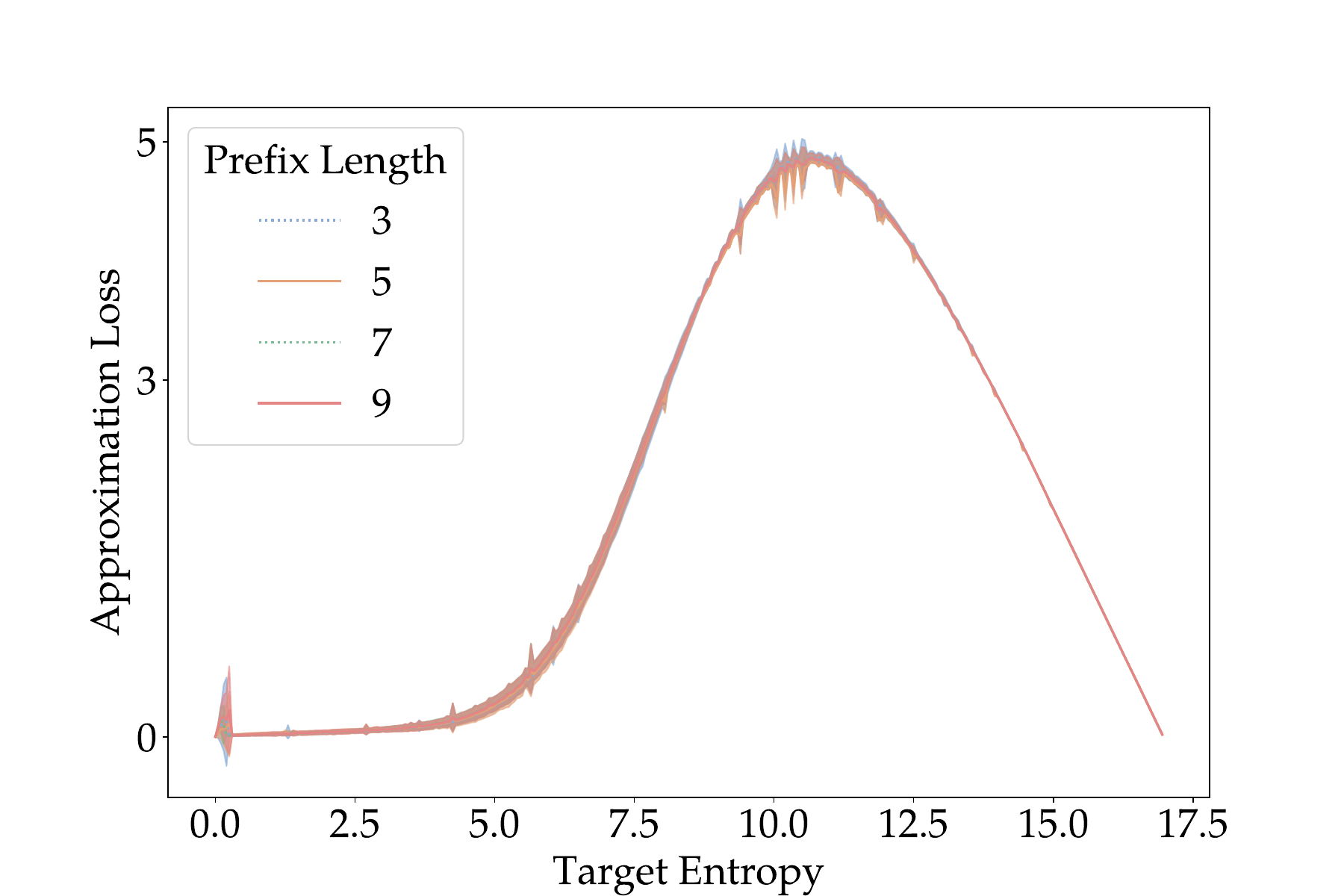}
    \caption{Approximation loss on vanilla distributions with varying prefix lengths. While different prefix lengths result in slight variations in approximation loss, the overall trend remains consistent.}
    \label{PrefixLength}
\end{figure}

We will gradually find a value $m$ that makes this inequality hold as equality, and then round it down to 0.01. In practice, we solve the above equation regarding different target probability distributions containing outliers with target entropy value $e$.

\section{Experiment Details}\label{ExperimentDetails}
The objective of our research is to determine whether, given a target distribution, it is possible to identify a prompt embedding that adequately approximates it. 
Since exhaustively enumerating all possible prompt embeddings is computationally infeasible, we instead employ modern optimization techniques, such as gradient descent, to search for an effective solution.

In our experiments, the prefix length is fixed at 5, and the learning rate is consistently set to 0.1 across all settings. The tuning process is conducted for a maximum of 500 epochs; however, early termination is applied if the approximation loss with respect to the target distribution does not exhibit improvement over 20 consecutive epochs. The minimum approximation loss observed throughout the training trajectory is recorded. The subsequent sections provide a detailed overview of the experimental setup.

\subsection{Prefix Initialization}

\citet{li_prefix-tuning_2021} observed that prefix initialization has a significant impact in low-data regimes. Given that our research aims to investigate the ease of eliciting specific target distributions using modern optimization techniques, it is crucial to minimize the influence of randomness introduced by different initialization configurations. To this end, we performed experiments using various initialization strategies by varying random seeds, thereby reducing the variability attributable to prompt embedding initialization.

We observed that the approximation loss varies across different initializations, as illustrated in \cref{Vanilla_Framework_Compare_Llama3}. In the main body of this study, we report the minimum approximation loss obtained from five different initializations. While the absolute approximation values differ depending on the initialization, we find that the overall trends observed across the studied distributions remain consistent.

\subsection{Prefix Length}

A longer prefix introduces more learnable parameters, which may potentially improve performance in approximating target distributions. To investigate this, we conducted an ablation study on prefix length using the vanilla distributions, as shown in \cref{Vanilla Distribution}. We found that increasing the number of tokens in the prefix—thus making it more expressive—does not alter the general trend. The results of approximating distributions with varying entropy under different prefix length configurations are presented in \cref{PrefixLength}.

\section{Future Directions}\label{FutureDirections}
We provide a case study investigating how well LMs approximate uniform distributions over tokens sharing the same part of speech.
This illustrative example demonstrates the applicability of our proposed framework (\cref{fig: frameworks}) for analyzing the expressiveness of LMs.
We envision that future work can further build on this framework to explore model behaviors from new interpretability perspectives.

\begin{figure}[t]
    \includegraphics[width=\columnwidth]{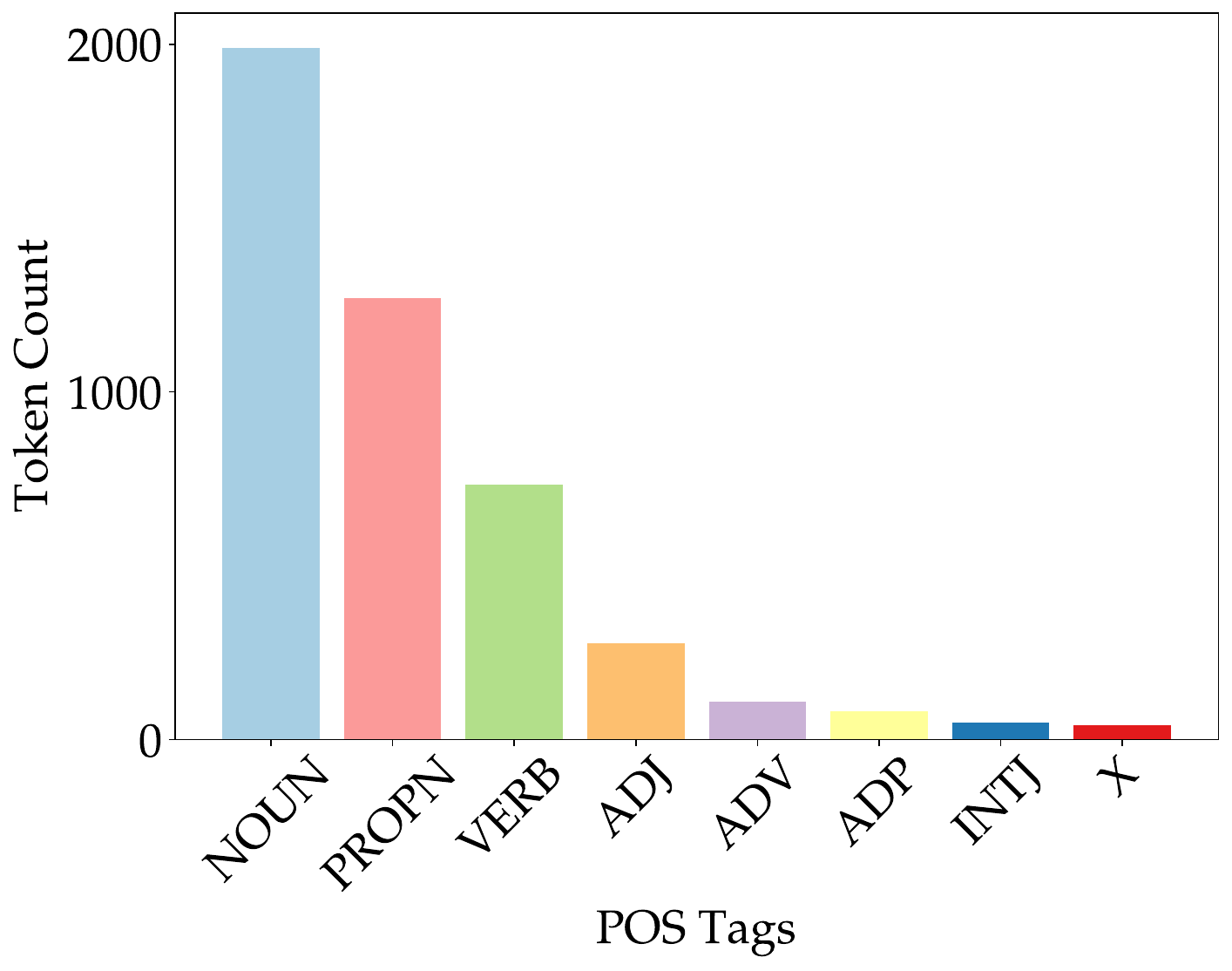}
    \caption{POS distribution of English tokens in the vocabularies of investigated LMs.}
    \label{DistributionPOS}
\end{figure}

We design a simple experiment to investigate the difficulty of approximating uniform distributions over tokens belonging to different parts of speech (POS).  
Specifically, we select three LMs from two distinct families and their instruction-tuned variants: Llama3.2-1B~\citep{dubey2024llama}, Qwen2.5-0.5B, and Qwen2.5-1.5B~\citep{DBLP:journals/corr/abs-2412-15115}.  
To identify the POS tags of tokens from the model vocabularies, we utilize ScispaCy~\citep{DBLP:conf/bionlp/NeumannKBA19} for tokenization and tagging.  
For simplicity, we restrict our analysis to tokens that are assigned a single, unambiguous POS tag.  
\cref{DistributionPOS} shows the distribution of the eight most common POS tags across English tokens in the vocabularies of the two model families.

We construct POS-wise uniform distributions by randomly sampling 100 tokens for nouns, proper nouns, verbs, and adjectives for 100 times, and create a uniform distribution each time.
Following the same setup in \cref{ExperimentDetails}, we report the minimum approximation loss across 5 different prompt initializations.

\begin{figure}[t!]
    \centering
    \begin{subfigure}[b]{0.45\textwidth}
        \centering
        \includegraphics[width=\textwidth]{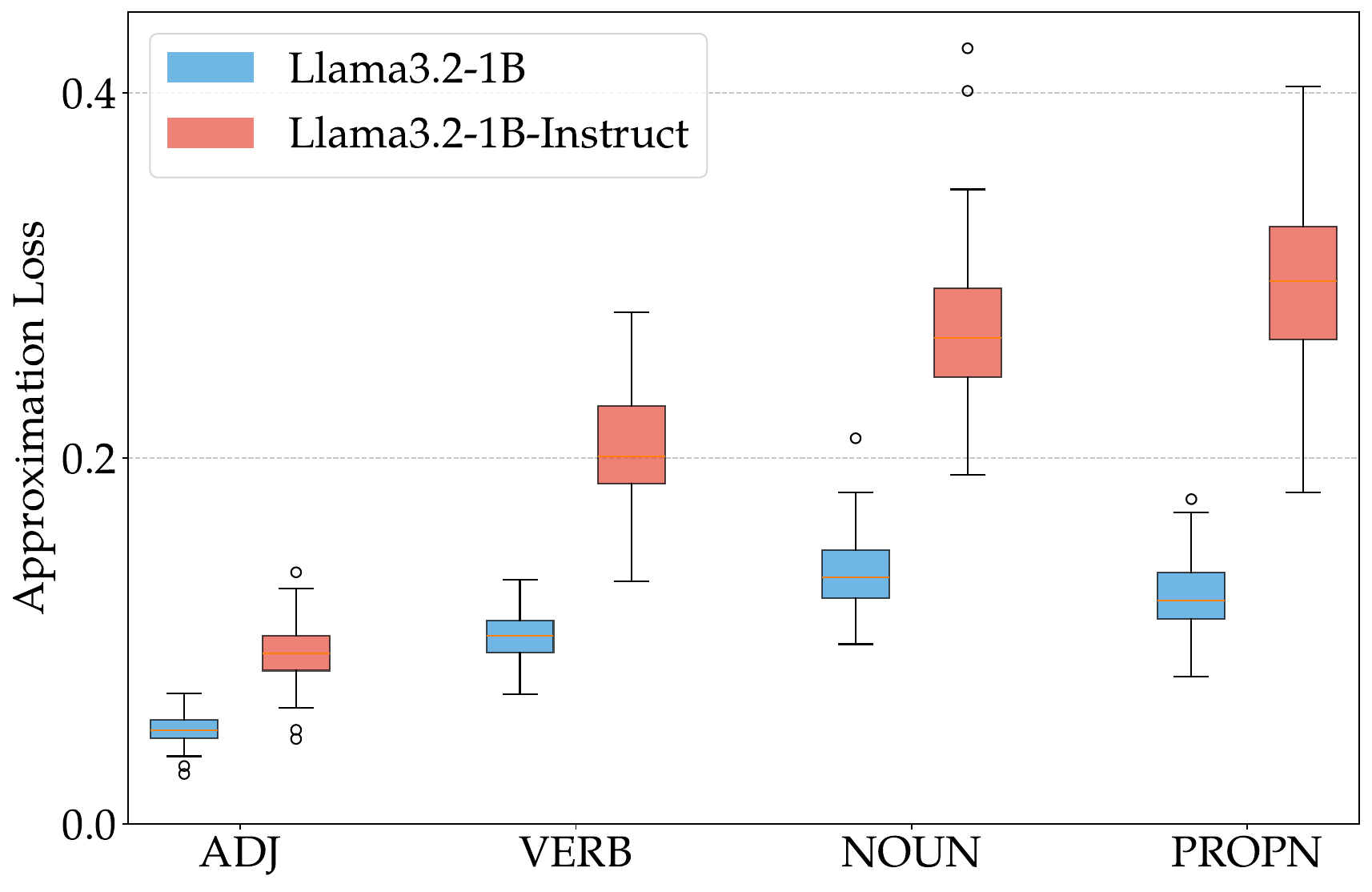}
        \subcaption{Llama3.2-1B}
        \label{Llama3_1B_instruct_base}
    \end{subfigure}
    \begin{subfigure}[b]{0.45\textwidth}
        \centering
        \includegraphics[width=\textwidth]{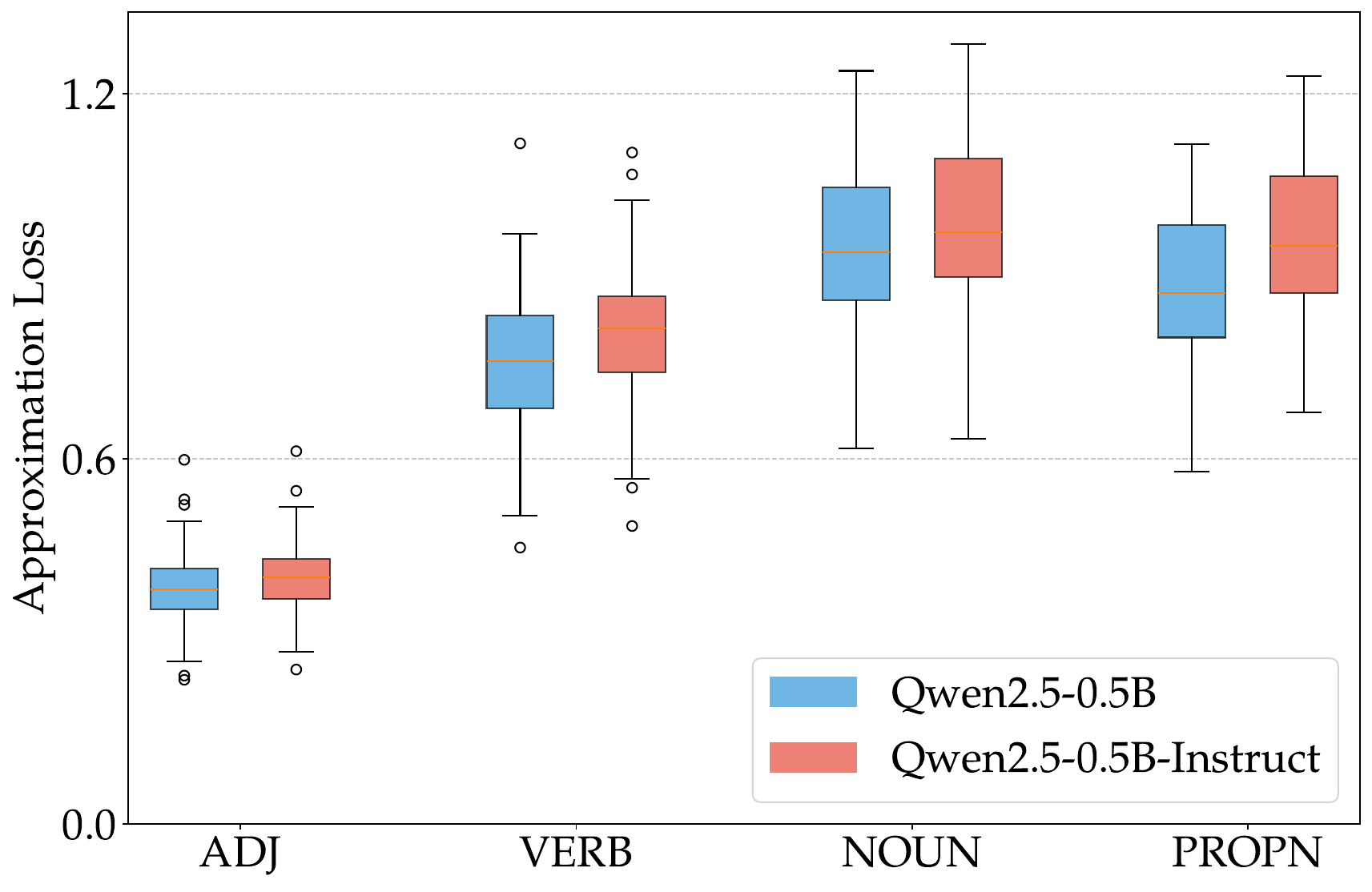}
        \subcaption{Qwen2.5-0.5B}
        \label{Qwen2_5_0_5_B_instruct_base}
    \end{subfigure}
    \begin{subfigure}[b]{0.45\textwidth}
        \centering
        \includegraphics[width=\textwidth]{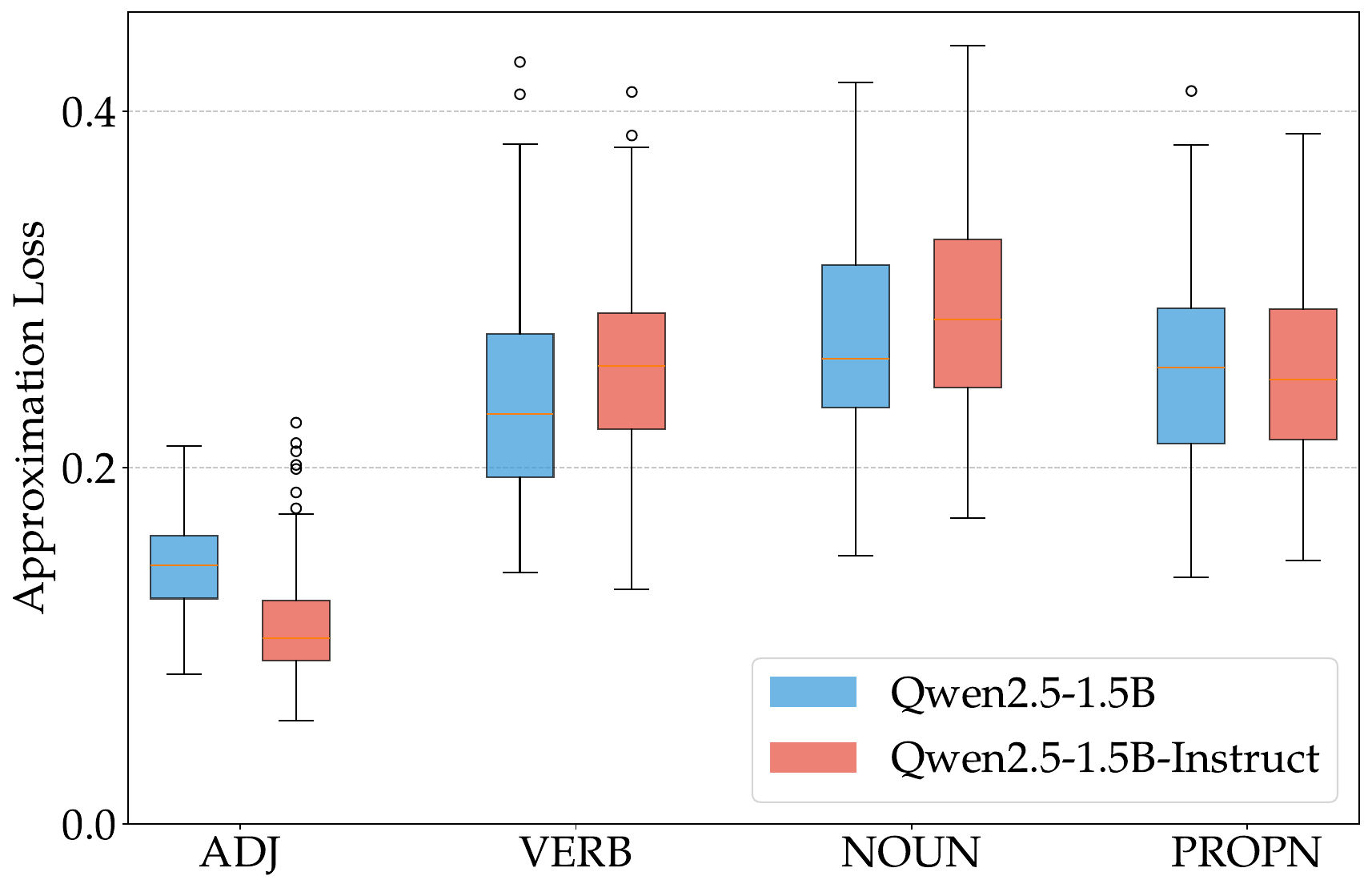}
        \subcaption{Qwen2.5-1.5B}
        \label{Qwen2_5_1_5_B_instruct_base}
    \end{subfigure}

    \caption{
The boxplot of approximation loss is plotted for POS-wise uniform distributions on LMs from two distinct families.
We report the four most common POS tags across English tokens in the vocabularies of the two model families, with comparison between base and instruction-tuned variants.
}
    \label{FutureWork}
\end{figure}

The experimental results are presented in \cref{FutureWork}.  
We observe that across all three comparisons, the instruction-tuned variants consistently exhibit higher approximation loss than their base counterparts.  
This suggests the presence of systematic distributional shifts introduced by instruction tuning.  
Additionally, we find that all evaluated LMs achieve lower approximation loss on adjectives compared to other POS.

These consistent patterns across models from different family suggest that applying our proposed framework from a distributional perspective offers a promising direction for systematically understanding LMs.
We hope this work inspires further research into the underlying distributional behaviors of LMs and informs future design of more robust and interpretable systems.

\end{document}